\documentclass[sigconf]{acmart}
\renewcommand\footnotetextcopyrightpermission[1]{} 
\usepackage{multirow}
\usepackage{bm}

\usepackage{bbding}
\usepackage{pifont}

\usepackage{subfig}
\usepackage{hhline}

\makeatletter
\@ifundefined{Bbbk}{}{}
\makeatother

\usepackage{amssymb}

\usepackage{booktabs}
\usepackage{tabularx}
\usepackage{array}
\usepackage{makecell}
\usepackage{pifont}
\usepackage{amsmath}

\usepackage{amsthm}
\usepackage{lineno}
\usepackage{textcomp}
\usepackage{xcolor}
\usepackage{enumitem}
\usepackage{color}
\usepackage{booktabs}
\usepackage{multirow}
\usepackage{makecell}
\usepackage{colortbl}
\usepackage{scalerel}
\usepackage{bigstrut}
\usepackage{bbding}
\usepackage{etoolbox}
\usepackage{newfloat}
\usepackage{listings}

\usepackage{algorithm}
\usepackage{algpseudocode}   

\begin{document}


\title{SEDformer: Event-Synchronous Spiking Transformers for \\ Irregular Telemetry Time Series Forecasting }

\author{
Ziyu Zhou\textsuperscript{\rm 1},
Yuchen Fang\textsuperscript{\rm 2},
Weilin Ruan\textsuperscript{\rm 1},
Shiyu Wang,
James Kwok\textsuperscript{\rm 3},
Yuxuan Liang\textsuperscript{\rm 1}
}
\authornote{Yuxuan Liang is the corresponding author.}

\affiliation{%
  \institution{
  \textsuperscript{\rm 1}The Hong Kong University of Science and Technology (Guangzhou) \\
  \textsuperscript{\rm 2}University of Electronic Science and Technology of China
  \textsuperscript{\rm 3}The Hong Kong University of Science and Technology}
  \city{}
  \state{}
  \country{}
}

\email{{ziyuzhou30,fyclmiss,rwlinno,kwuking}@gmail.com,
jamesk@cse.ust.hk, yuxliang@outlook.com}

\begin{abstract}
Telemetry streams from large-scale Internet-connected systems (e.g., IoT deployments and online platforms) naturally form an irregular multivariate time series (IMTS) whose accurate forecasting is operationally vital. A closer examination reveals a defining Sparsity–Event Duality (SED) property of IMTS, i.e., long stretches with sparse or no observations are punctuated by short, dense bursts where most semantic events (observations) occur. However, existing Graph- and Transformer-based forecasters ignore SED: pre-alignment to uniform grids with heavy padding violates sparsity by inflating sequences and forcing computation at non-informative steps, while relational recasting weakens event semantics by disrupting local temporal continuity. These limitations motivate a more faithful and natural modeling paradigm for IMTS that aligns with its SED property. We find that Spiking Neural Networks meet this requirement, as they communicate via sparse binary spikes and update in an event-driven manner, aligning naturally with the SED nature of IMTS. Therefore, we present \textbf{SEDformer}, an \textbf{SED}-enhanced Spiking Trans\textbf{former} for telemetry IMTS forecasting that couples: (1) a \textbf{SED-based Spike Encoder} converts raw observations into event synchronous spikes using an \textbf{Event-Aligned LIF} neuron, (2) an \textbf{Event-Preserving Temporal Downsampling} module compresses long gaps while retaining salient firings and (3) a stack of \textbf{SED-based Spike Transformer} blocks enable intra-series dependency modeling with a membrane-based linear attention driven by EA-LIF spiking features. Experiments on public telemetry IMTS datasets show that SEDformer attains state-of-the-art forecasting accuracy while reducing energy and memory usage, providing a natural and efficient path for modeling IMTS.
\end{abstract}

\maketitle

\vspace{-2mm}
\section{Introduction}
\vspace{-1mm}
Time series are ubiquitous in large-scale Internet-connected systems, including IoT deployments and online platforms: sensor readings, device logs, service metrics, and user-facing workload signals all evolve over time and are routinely forecast to allocate resources, protect service-level agreements, and personalize user experiences \cite{hou2022multi,jhin2022exit,jiang2023learning,miyake2024netevolve,ning2024debiasing,TSFool,SDformer}. However, in operational settings such telemetry is often irregularly sampled because activity dynamics generate bursts and lulls and measurement systems introduce uneven logging delays \cite{crovella2002self,akidau2013millwheel,akidau2015dataflow}. Together these factors yield observations that arrive at non-uniform intervals and are asynchronous across variates, resulting in an irregular multivariate time series (IMTS) \cite{Che2018,GRU-ODE}. Accurate telemetry IMTS forecasting is thus pivotal for capacity planning \cite{almeida2002capacity}, anomaly response \cite{ren2019time,xie2025multivariate}, traffic shaping \cite{jain2013b4}, and recommendation updates \cite{covington2016deep} in large-scale Internet-connected systems.

\begin{figure}[t]
    \centering
    \includegraphics[width=0.95\linewidth, trim=20 15 20 30, clip]{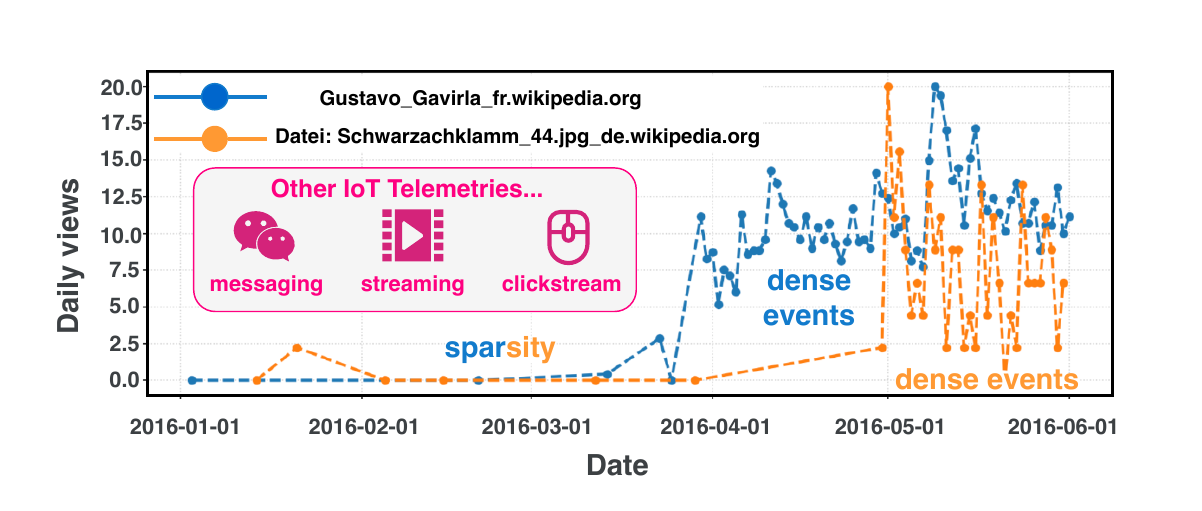}
    \vspace{-3mm}
    \caption{Five-month daily Wikipedia page views for two articles, exhibiting both sparsity and dense events. For brevity, we abbreviate the titles as ``Gustavo'' and ``Datei''. Beyond article page views, telemetry IMTS also arises in IoT sensing, messaging, video streaming, and clickstream, where sparsity--event duality similarly holds.}
    \label{fig:intro}
    \vspace{-6mm}
\end{figure}

A telemetry IMTS typically consists of long stretches with few or no observations (i.e., sparse sampling), interleaved with short, dense bursts during which most semantically meaningful observations (events) occur. Fig. \ref{fig:intro} illustrates this pattern with two Wikipedia page-view series \cite{WikeArticle}. ``Gustavo'' is extremely sparse from January to April, then exhibits many closely-spaced observations starting in April, while ``Datei'' remains sparse and spikes in May. We call this property \textbf{Sparsity–Event Duality (SED)}. Preserving event semantics while simultaneously exploiting sparsity for efficiency is a natural and effective way to model IMTS.

Deep learning 
models 
for IMTS forecasting 
are typically based 
on either
Ordinary Differential Equations \cite{Neural-ODE, Latent-ODE,GRU-ODE}, Graph Neural Networks~\cite{RainDrop,tPatchGNN,TimeCHEAT}, or Transformers \cite{Warpformer,ContiFormer}. While 
effective, 
they
do not make use of the Sparsity-Event Duality. First, 
the IMTS 
is simply aligned 
to a regular 
series-level or patch-level
grid,
with
the unobserved values
replaced by zeroes
\cite{Che2018,mTAND,Neural-Flows,Warpformer,tPatchGNN,liu2025rethinkingirregulartimeseries}. This results in unnecessary computations and an overly-complex model. Moreover, as all time steps (including the padded zeroes) are treated equally, this can undermine the event semantics
\cite{Warpformer,tPatchGNN,liu2025rethinkingirregulartimeseries}. Other approaches
transform the IMTS to a relational structure such as point set \cite{lin2024sparsetsf} or bipartite graph \cite{GraFITi,TimeCHEAT}. However, this also disrupts the continuity of local events in the time series, and the fundamental event-driven nature of IMTS is diluted or even destroyed.

Motivated by these observations,
we raise the question: {\em Is there a \textbf{naturally appropriate} architecture for IMTS forecasting?} We posit that Spiking Neural Networks (SNNs) \cite{maass1997networks} might be a strong candidate. First, SNNs communicate via sparse, discrete spike trains, which are activated only when informative stimuli occur \cite{zhou2022spikformerspikingneuralnetwork,AttentionSpikingNeuralNetworks}. This leads to substantial savings in computation and energy, and aligns with the sparse nature of IMTS where information is limited and temporally localized. Second, SNNs follow an event-driven computation paradigm \cite{yao2023spikedriventransformer,wu2025spikf}: neurons update and emit spikes only upon events rather than at every tick of a fixed grid. This matches the event-driven nature of IMTS, where salient semantics concentrate within short, dense bursts of observations.

In this paper, we propose the \textbf{S}parsity–\textbf{E}vent \textbf{D}uality enhanced Spiking Trans\textbf{former} (\textbf{SEDformer}) for telemetry IMTS forecasting. We operationalize the SED of IMTS with a four-stage, end-to-end design. First, to exploit the strength of spiking neurons on sparse, event-driven IMTS, we first align the time steps between IMTS and the SNN through \textbf{SED-based Spike Encoder (SED-SE)} where the model updates only when an observation (event) arrives. At its core is an Event-Aligned LIF (EA-LIF) neuron whose leak depends on the inter-event interval, making the representation explicitly interval-aware so that long gaps decay and informative bursts dominate. Because telemetry streams contain many long idle stretches that offer little value, the \textbf{Event-Preserving Temporal Downsampling (EPTD)} module then pools along the event axis to retain whether bursts occurred while compressing extended gaps, reducing compute without reintroducing a uniform grid. On the condensed event-synchronous sequence, \textbf{SED-based Spike Transformer (SED-ST)} applies interval-conditioned, membrane-based linear attention whose scores depend on inter-event gaps, emphasizing bursts and down-weighting long silences while keeping complexity linear in the number of pooled events. Finally, a lightweight MLP-based \textbf{Query-Aware Decoder} enables query‑specific forecasts. The entire architecture is trained end-to-end with mean-squared error (MSE). To the best of our knowledge, SEDformer is the first SNN-based framework for IMTS forecasting, rethinking IMTS forecasting from its defining SED property, preserving sparsity and event semantics while capturing intra-series temporal dependencies and maintaining high computational efficiency. Our contributions can be summarized as follows:
\begin{itemize}[leftmargin=*]
    \item We identify the Sparsity-Event Duality (SED) of telemetry IMTS, arguing that preserving event semantics while exploiting sparsity is crucial for effective forecasting.
    \item We introduce SEDformer, which (i) aligns computation to observed events and injects interval information via the newly proposed EA-LIF neuron, (ii) compresses long gaps while preserving salient firings through max pooling, and (iii) learns intra-series temporal dependencies with a membrane-driven, irregularity-conditioned self-attention mechanism for event-aware and efficient representation learning.
    \item Extensive experiments on public telemetry IMTS datasets show that SEDformer 
    outperforms 
    the state-of-the-art in terms of both
    efficiency and
    accuracy. 
\end{itemize}

\vspace{-5mm}

\section{Related Works}
\vspace{-1mm}
\subsection{Telemetry \& IMTS Forecasting}
\vspace{-1mm}
\label{sec:relatedworks}
Telemetry forecasting spans IoT sensing, device/service metrics, and workload signals such as page views and query volumes. Classical and industry tools such as Prophet emphasize decomposable trend and seasonality for large deployments \cite{taylor2018forecasting}, while early large-scale demand signals were modeled from search and query logs \cite{choi2012predicting}. Popularity and traffic bursts of online content have been anticipated using early-signal dynamics and log-derived features \cite{szabo2010predicting,pinto2013using}. Deep learning has since become dominant: DeepAR delivers probabilistic forecasts with autoregressive RNNs at scale \cite{salinas2020deepar}, N-BEATS advances univariate accuracy with a pure MLP stack \cite{oreshkin2019n}, and Temporal Fusion Transformers fuse attention with multi-horizon interpretability \cite{lim2021temporal}. Yet, these lines largely overlook irregular telemetry and do not explicitly model non-uniform sampling. From a broader IMTS perspective, continuous-time models parameterize temporal evolution via Differential Equation (DE): Neural ODE \cite{Neural-ODE}, Latent ODE \cite{Latent-ODE}, CRU \cite{CRU}, and GRU-ODE \cite{GRU-ODE} with Neural Flows avoiding costly solvers \cite{Neural-Flows}, and ContiFormer integrating ODE dynamics with attention to report gains on irregular data \cite{ContiFormer}. Complementary relational formulations recast IMTS as graphs or sets: GraFITi uses sparse bipartite graphs \cite{GraFITi}, tPatchGNN learns transformable patches with inter-patch GNNs \cite{tPatchGNN}, Hi-Patch aggregates multi-scale dense variates \cite{Hi-Patch}, and HyperIMTS reasons over hypergraph structures \cite{HyperIMTS}. However, as summarized in Tab.~\ref{tab:related-work}, representative methods still face limitations: DE and Transformer hybrids can be computation heavy when numerical solvers are involved, while padding-based pre-alignment to uniform grids introduces substantial zeros that inflate sequence length (violating sparsity) and blur the local event semantics central to IMTS.

\begin{table}[t]
\vspace{-2mm}
\caption{Comparison between prior IMTS forecasters and our proposed SEDformer.}
\label{tab:related-work}
\vspace{-4mm}
\centering
\scriptsize
\setlength{\tabcolsep}{2pt}
\renewcommand{\arraystretch}{1.0}

\begin{tabularx}{\columnwidth}{
  >{\centering\arraybackslash}p{0.23\columnwidth}
  >{\centering\arraybackslash}X
  >{\centering\arraybackslash}X
  >{\centering\arraybackslash}X
}
\toprule[1.2pt]
\makecell{\textbf{IMTS}\\\textbf{Forecaster}} &
\makecell{\textbf{Sparsity}\\\textbf{-Preserving}} &
\makecell{\textbf{Event}\\\textbf{-Aligning}} &
\makecell{\textbf{Energy/Cost}\\\textbf{Saving}} \\
\midrule
GraFITi~\cite{GraFITi}         & \ding{51} & \ding{55} & \ding{55} \\
tPatchGNN~\cite{tPatchGNN}     & \ding{55} & \ding{55} & \ding{51} \\
TimeCHEAT~\cite{TimeCHEAT}     & \ding{55} & \ding{55} & \ding{55} \\
HyperIMTS~\cite{HyperIMTS}     & \ding{51} & \ding{55} & \ding{51} \\
KAFNet~\cite{kafnet}           & \ding{55} & \ding{55} & \ding{51} \\
APN~\cite{liu2025rethinkingirregulartimeseries} & \ding{55} & \ding{55} & \ding{51} \\
\textbf{SEDformer}             & \ding{51} & \ding{51} & \ding{51} \\
\bottomrule[1.2pt]
\end{tabularx}
\vspace{-6mm}
\end{table}

\vspace{-2mm}

\begin{figure*}[t]
    \centering
    \includegraphics[width=0.82\linewidth, trim=18 13 20 10, clip]{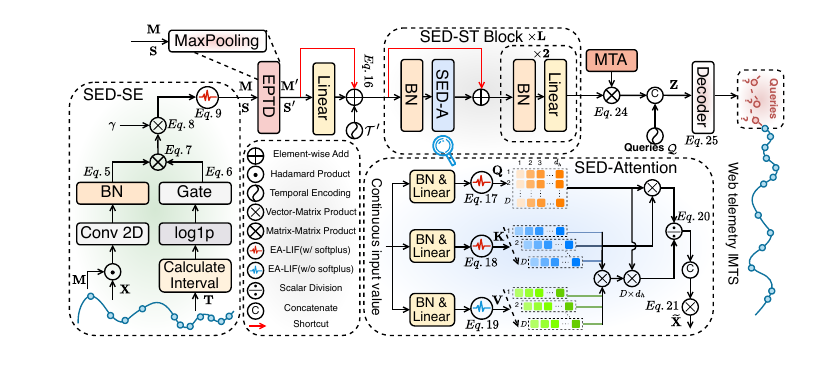}
    \vspace{-3mm}
    \caption{The architecture of SEDformer. SED-based Spike Encoder (SED-SE) initially converts IMTS into event-synchronous spike trains via an Event-Aligned LIF (EA-LIF) neuron. Event-Preserving Temporal Downsampling (EPTD) module then retains salient spikes while collapsing long gaps. A stack of SED-based Spike Transformer (SED-ST) blocks subsequently learns temporal dependencies via membrane-based SED-Attention mechanism, and Masked Time Aggregation (MTA) summarizes each variate over observed events. A lightweight Decoder finally maps the summaries to forecasts at future times.}
    \label{fig:SEDformer}
    \vspace{-3mm}
\end{figure*}

\subsection{SNNs for Temporal Dynamics Modeling}
\vspace{-1mm}
Early applications of Spiking Neural Networks (SNNs) to time series,
as exemplified by the NeuCube framework for spatio-temporal prediction \cite{KASABOV201462},
relied on hand-crafted encoders or reservoir-style dynamics.
With the advent of surrogate-gradient training \cite{wu2019direct}, trainable spiking architectures are developed for sequential modeling. Examples include Spike-TCN \cite{lv2024efficient}, Spike-RNN \cite{lv2024efficient}, iSpikformer \cite{lv2024efficient} and SpikF \cite{wu2025spikf}, which demonstrate the feasibility of combining SNNs with attention and Fourier representations. In parallel, researchers have sought to enhance the expressivity of spiking neurons. Several works revise the vanilla Leaky
Integrate-and-Fire (LIF) neuron by incorporating richer temporal dynamics, such as STC-LIF \cite{wang2024autaptic} and TS-LIF \cite{TS-LIF}, which aim to better capture short-term correlations and long-range dependencies. Along this line, the Learnable Multi-hierarchical (LM-H) neuron introduces learnable parameters to adaptively balance the historical and current information \cite{hao2023progressive}, and the multi-compartment neuron model leverages temporal dendritic heterogeneity to process multi-scale inputs \cite{zheng2024temporal}. Despite recent advances, most SNN-based forecasting models still assume regular sampling, leaving their application to IMTS largely underexplored. They process fixed-step sequences and do not encode inter-event intervals in neuron or attention dynamics. As a result, they overlook the irregularity typical in IMTS and compromise its Sparsity–Event Duality (SED).

\vspace{-2mm}

\section{Preliminary}
\label{sec:preliminary}

\vspace{-1mm}
\subsection{Definition and Problem Formulation}
\label{subsec:problem_formulation}
\vspace{-1mm}
\subsubsection*{Definition (IMTS and Events).}
We denote
an irregular multivariate time series (IMTS) with $D$ variates as
\(\mathcal{O}=\{[(t_i^{d},x_i^{d})]_{i=1}^{L_d}\}_{d=1}^{D}\), where 
each variate $d$ has 
\(L_d\) 
\emph{events}. Each event
\((t_i^{d},x_i^{d})\in\mathbb{R}^2\) 
is measured at time \(t_i^{d}\) and has value \(x_i^{d}\). 
The sampling intervals in a variate are generally non-uniform (intra-variate irregularity), and event times across variates are not synchronized (inter-variate asynchrony). To obtain a global timeline, we take the union 
\(
\bigcup_{d=1}^{D}\{t^{d}_{1},\dots,t^{d}_{L_d}\}
\)
and
sort them as
\(\mathcal{T}=\{t_1\le\cdots\le t_K\}.\)
The observations 
are aligned along
this axis to form matrices
\(\mathbf{T},\mathbf{X},\mathbf{M}\in\mathbb{R}^{K\times D}\), where \(\mathbf{T}[k]=t_k\), 
\(\mathbf{X}[k,d]\) stores the value of variate \(d\) at event \(t_k\) if observed and
\(\mathbf{M}[k,d]\in\{0,1\}\) indicates that the measurement is observed at time $t_k$ (\(\mathbf{M}[k,d]=1\)) or not
(\(\mathbf{M}[k,d]=0\)).
We refer to
\((\mathbf{T},\mathbf{X},\mathbf{M})\) as the event-aligned representation.
\vspace{-2mm}

\subsubsection*{Problem (IMTS Forecasting).}
Let \(\mathcal{Q}=\{[q_j^{d}]_{j=1}^{Q_d}\}_{d=1}^{D}\) be the matrix containing 
timestamps
for future queries
(with \(q_j^{d}>\max_{1\le i\le L_d} t_i^{d}\))
of all the $D$ variates.
The goal is to learn a model \(\mathcal{F}_\theta\) that maps historical observations 
$\mathcal{O}$
and queries 
$\mathcal{Q}$
to predictions
\(
\hat{\mathcal{X}}=\{[\hat{x}_j^{d}]_{j=1}^{Q_d}\}_{d=1}^{D},
\)
where \(\hat{x}_j^{d}\) is the corresponding prediction 
at time \(q_j^{d}\)
for variate \(d\).

\subsection{Vanilla LIF Neuron and Surrogate Gradient}
\label{sec:prelim_lif}

The Leaky Integrate–and–Fire (LIF) \cite{maass1997networks} neuron is the canonical spiking unit in SNNs, striking a balance between biological plausibility and computational simplicity~\cite{gerstner2014neuronal}. Its discrete–time dynamics comprise \emph{integration}, \emph{leakage}, \emph{thresholding}, and \emph{reset}. Specifically, at layer $l$, let $\boldsymbol{s}^{\,l-1}[t]\in\{0,1\}$ be the spike output from layer $l\!-\!1$:
\begin{eqnarray}
(\text{integration}) 
& \boldsymbol{x}^{l}[t] &= \boldsymbol{W}^{l}\,\boldsymbol{s}^{\,l-1}[t], 
\label{eq:lif-x}\\
(\text{leakage}) 
& \boldsymbol{m}^{l}[t] &= \alpha^{l}\,\boldsymbol{v}^{l}[t-1] + \bigl(1-\alpha^{l}\bigr)\,\boldsymbol{x}^{l}[t],
\label{eq:lif-m}\\
(\text{thresholding})
& \boldsymbol{s}^{l}[t] &= H\bigl(\boldsymbol{m}^{l}[t]-v_{\mathrm{th}}\bigr),
\label{eq:lif-s}\\
(\text{reset})
& \boldsymbol{v}^{l}[t] &= \boldsymbol{m}^{l}[t]-v_{\mathrm{th}}\,\boldsymbol{s}^{l}[t],
\label{eq:lif-v}
\end{eqnarray}
where $\boldsymbol{x}^{l}[t]$ is the synaptic current induced by $\boldsymbol{W}^{l}$, $\boldsymbol{m}^{l}[t]$ and $\boldsymbol{v}^{l}[t]$ are the pre- and post-spike membrane potentials, respectively, and $\alpha^{l}\in[0,1)$ is the leak coefficient controlling exponential decay (setting $\alpha^{l}=0$ recovers the non-leaky IF neuron). The Heaviside step function is $H(u)=\mathbf{1}\{u\ge 0\}$. This recurrence realizes exponentially weighted integration with leakage, acting as a low-pass memory with reset upon threshold crossing~\cite{wang2024autapticsynapticcircuitenhances}.

As gradients through $H(\cdot)$ are undefined, surrogate gradients are used for training~\cite{neftci2019surrogate,wu2019direct,lv2024efficient}. We adopt a straight–through sigmoid estimator (STE–sigmoid)~\cite{zenke2018superspike} and approximate
\(
\frac{\partial H(u)}{\partial u}\approx\alpha_{\mathrm{ste}}
\sigma(\alpha_{\mathrm{ste}} u)\bigl(1-\sigma(\alpha_{\mathrm{ste}} u)\bigr),
\) where $\sigma(\cdot)$ is the logistic sigmoid and $\alpha_{\mathrm{ste}}>0$ controls the slope. This enables end–to–end BPTT while leaving the forward LIF dynamics unchanged.

\vspace{-1mm}

\section{Methodology}
The SEDformer framework is shown in Fig. \ref{fig:SEDformer}. First, the SED-based Spike Encoder (SED-SE) converts IMTS into an event-aligned spike train via an Event-Aligned LIF (EA-LIF) with updates only on the observed events. The Event-Preserving Temporal Downsampling (EPTD) module then performs a temporal max-pooling on the spike timeline. It preserves salient firing events while compressing long inter-event gaps. Next, a stack of SED-based Spike Transformer (SED-ST) blocks
enable deeper representation learning of intra-series temporal irregularity.  Its underlying
SED-based self-attention mechanism leverages the EA-LIF–derived features conditioned on inter-event intervals.
Finally, the Query-Aware Decoder addresses arbitrary forecasting queries.

\vspace{-1mm}

\subsection{SED-based Spike Encoder (SED-SE)}
\label{sec:SED-SE}
 
SNNs are naturally event-driven and sparse, making them a promising fit for IMTS. To fully exploit this property, we align the temporal axis of SNN computation with the observed events of IMTS, rather than forcing signals onto a uniform grid. Prior work for regular series aligns an SNN step with a series step by expanding each step into multiple sub-steps (e.g., Delta spike encoders~\cite{delta-encoder}, convolutional spike encoders~\cite{lv2024efficient}, and ZOH upsampling~\cite{wu2025spikf}). These schemes assume or impose uniform sampling and thus cannot exploit irregular inter-event intervals. This motivates the proposed \textbf{SED-based Spike Encoder (SED-SE)}, which converts raw IMTS into event-synchronous sparse spike trains, preserving event semantics for downstream SNN processing.

\subsubsection{SED-based Spike Encoding Process.}
Let $\{t_k\}_{k=1}^{K}$ be the non-decreasing event times (the shared index from Sec.~\ref{subsec:problem_formulation}) with inter-event intervals $\Delta t_k=t_k-t_{k-1}\ge0$ and $\Delta t_1=0$, and let $\mathbf{X}\in\mathbb{R}^{K\times D}$ masked by $\mathbf{M}\in\mathbb{R}^{K\times D}$ (\(K\) is the number of events and \(D\) is the number of variates) through elementwise product $\odot$. Concretely, we reuse the event-aligned representation $(\mathbf{T},\mathbf{X},\mathbf{M})$ defined in Sec.~\ref{subsec:problem_formulation}, where $\mathbf{T}[k]=t_k$, $\mathbf{X}[k,d]$ stores the observed value of variate $d$ at event $t_k$ (if any), and $\mathbf{M}[k,d]\in\{0,1\}$ indicates whether that value is observed (1) or missing (0). We feed the masked input $\mathbf{X}\odot\mathbf{M}$ to ensure that unobserved entries are not encoded.
For each variate, we first extract $C$ local channels per event to capture short-range shape around the observation while avoiding cross-variate mixing. A depthwise $1\times k$ convolution over the masked sequence $(\mathbf{X}\odot\mathbf{M})$ aggregates a small temporal neighborhood centered at $t_k$, and batch normalization stabilizes the feature scale across events:\footnote{$\mathrm{Conv}_{1\times k}$ is depthwise across variates; $k$ is a small odd kernel with receptive field $\{t_{k-r},\dots,t_{k+r}\}$, $r=(k-1)/2$. $\mathrm{BN}(\cdot)$ denotes per-channel batch normalization.}
\begin{equation}
\label{eq:sedse-local-bn}
\boldsymbol{x}^{\,l}[t_k]
\;=\;
\mathrm{BN}\Big(
\mathrm{Conv}_{1\times k}\big((\mathbf{X}\odot\mathbf{M})[t_{\,k-r:k+r}]\big)\Big)
\;\in\;\mathbb{R}^{D\times C}.
\end{equation}
we subsequently inject interval information to reflect recency: longer gaps since the previous event should down-weight stale evidence, while short gaps should emphasize fresh observations. We encode this with a log-stabilized scalar gate per event:
\begin{equation}
\label{eq:sedse-gate-core}
\widehat{\Delta t}_k \;=\; \log\!\Big(1+\frac{\Delta t_k}{\rho^{l}}\Big),
\qquad
s_k \;=\; \sigma\big(a^{l}\,\widehat{\Delta t}_k + b^{l}\big),
\end{equation}
where $\rho^{l}\!>\!0$ is a learnable time scale and $a^{l},b^{l}\!\in\!\mathbb{R}$ are learnable gate parameters. The scalar $s_k\!\in\!(0,1)$ summarizes how recent the current event is on a normalized timeline. We then broadcast $s_k$ over variates and channels to modulate the local features:
\begin{equation}
\label{eq:sedse-gate-apply}
\tilde{\boldsymbol{x}}^{\,l}[t_k]
\;=\;
s_k\,\boldsymbol{x}^{\,l}[t_k],
\qquad
\boldsymbol{x}^{\,l}[t_k],\,\tilde{\boldsymbol{x}}^{\,l}[t_k]\in\mathbb{R}^{D\times C},
\end{equation}
so that interval-aware evidence is passed downstream while preserving the event-synchronous layout.
Eq.~\ref{eq:sedse-gate-apply} injects interval awareness into the feature map while preserving its shape for downstream EA-LIF processing. Then, we center and scale the gated features to obtain an event–aligned synaptic current:
\begin{equation}
\label{eq:sedse-current}
\boldsymbol{I}^{\,l}[t_k]
\;=\;
\gamma^{l}\!\big(\,\tilde{\boldsymbol{x}}^{\,l}[t_k]-\theta^{l}\,\big),
\qquad
\boldsymbol{I}^{\,l}[t_k]\in\mathbb{R}^{D\times C}.
\end{equation}
Here $\gamma^{l}\!>\!0$ is a learnable gain (scalar or per–channel, shared across events), and $\theta^{l}\!\in\!\mathbb{R}^{C}$ is a learnable per–channel offset that is broadcast along the variate dimension $D$. This operation produces a zero–referenced, amplitude–controlled current at the same event grid $\{t_k\}$. Finally, we convert these event-aligned currents into spike trains by applying the Event-Aligned LIF (EA-LIF, Sec.~\ref{sec:ea-lif}) neuron on the same event grid \(t_k\):
\begin{equation}
\label{eq:sedse-to-spike}
\resizebox{0.9\linewidth}{!}{$
\bigl(\boldsymbol{m}^{l}[t_k],\boldsymbol{s}^{l}[t_k],\boldsymbol{v}^{l}[t_k]\bigr)
=
\mathcal{EA\mbox{-}LIF}\Big(\boldsymbol{v}^{l}[t_{k-1}],\boldsymbol{I}^{\,l}[t_k];\,\Delta t_k,\,\tau^{l},\,v_{\mathrm{th}}\Big)
$}
\end{equation}
Here, \(\boldsymbol{v}^{l}[t_{k-1}]\) denotes the previous post-spike membrane potential, \(\boldsymbol{I}^{\,l}[t_k]\) the event-aligned synaptic current, \(\Delta t_k\) the inter-event interval, \(\tau^{l}\) the time constant, and \(v_{\mathrm{th}}\) the firing threshold. This produces spike trains as the output of SED-SE:
\(
\mathbf{S}
\;=\;
\big\{\boldsymbol{s}^{l}[t_k]\big\}_{k=1}^{K}
\;\in\;
\mathbb{R}^{K\times D\times C}.
\)
Practically, we use a straight–through
sigmoid estimator (STE–sigmoid) to backpropagate through the binary firing in EA-LIF. In sum, SED-SE delivers an event-synchronous spike-encoding scheme that captures temporal variation and better represents the dynamic nature of IMTS.

\begin{figure}[t]
    \centering
    \includegraphics[width=\linewidth, trim=20 15 20 15, clip]{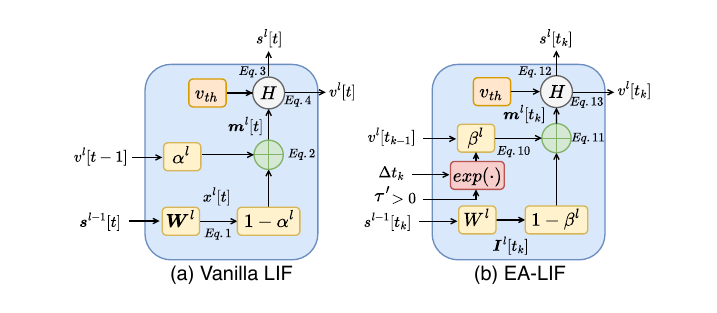}
    \vspace{-5mm}
    \caption{The structure of the (a) vanilla LIF and the (b) Event-Aligned LIF (EA-LIF) we propose.}
    \label{fig:EALIF}
    \vspace{-5mm}
\end{figure}

\subsubsection{Event-Aligned LIF (EA-LIF)}
\label{sec:ea-lif}
The vanilla LIF neuron in Sec.~\ref{sec:prelim_lif} (Eqs.~\ref{eq:lif-x}–\ref{eq:lif-v}) uses a \textbf{constant} leak $\alpha^{l}$ per step, which implicitly presumes a fixed sampling interval and thus cannot reflect variable gaps between observations. To accommodate IMTS, we replace the constant leak with an interval-dependent decay that matches the continuous LIF solution: if a membrane potential follows $\frac{d v}{dt}=-\frac{1}{\tau^{l}}v$ between two events separated by $\Delta t_k$, then $v$ decays multiplicatively by $\exp(-\Delta t_k/\tau^{l})$. Hence we define an event-dependent leak:
\begin{equation}
\label{eq:ease-beta}
\beta^{l}[t_k] \;=\; \exp\!\Big(-\frac{\Delta t_k}{\tau^{l}}\Big), \qquad \tau^{l}>0 .
\end{equation}
To ensure $\tau^{l}$ remains positive and trainable, we parameterize it as
$\tau^{l}=\mathrm{softplus}(\eta^{l})+1$, where $\mathrm{softplus}(u)=\log(1+e^{u})$. Given the event-aligned current $\boldsymbol{I}^{\,l}[t_k]$ from Eq.~\ref{eq:sedse-current}, EA-LIF updates only at event times $\{t_k\}$:
\begin{equation}
\label{eq:ease-m}
\boldsymbol{m}^{l}[t_k] \;=\; \beta^{l}[t_k]\;\boldsymbol{v}^{l}[t_{k-1}] \;+\; \bigl(1-\beta^{l}[t_k]\bigr)\,\boldsymbol{I}^{\,l}[t_k],
\end{equation}
\begin{equation}
\label{eq:ease-s}
\boldsymbol{s}^{l}[t_k] \;=\; H\bigl(\boldsymbol{m}^{l}[t_k]-v_{\mathrm{th}}\bigr),
\end{equation}
\begin{equation}
\label{eq:ease-v}
\boldsymbol{v}^{l}[t_k] \;=\; \boldsymbol{m}^{l}[t_k]-v_{\mathrm{th}}\,\boldsymbol{s}^{l}[t_k].
\end{equation}
Conceptually, Eqs.~\ref{eq:ease-s}–\ref{eq:ease-v} have the same thresholding and reset forms as Eqs.~\ref{eq:lif-s}–\ref{eq:lif-v} in vanilla LIF. The key difference is that the decay factor \(\beta^{l}[t_k]\) now depends on the actual inter-event gap \(\Delta t_k\), and all updates are indexed on the event grid \(\{t_k\}\) rather than a uniform step index \(t\). EA-LIF is therefore naturally aligned with IMTS: computation occurs precisely at observed events (preserving event semantics), while \(\beta^{l}[t_k]\) adapts the leakage to the elapsed time. Therefore, the state decays strongly after long silences and retains memory within dense bursts, respecting sparsity without inserting artificial time steps. It is worth noting in the special case of regular sampling (\(\Delta t_k\!\equiv\!\Delta\)), \(\beta^{l}[t_k]\!\equiv\!e^{-\Delta/\tau^{l}}\) becomes constant, and EA-LIF reduces to vanilla LIF by identifying \(\alpha^{l}=\beta^{l}[t_k]\).

\subsection{Event-Preserving Temporal Downsampling}
\label{sec:downsample}

Following SED-SE, we obtain event-synchronous spikes
\(\mathbf{S}\!\in\!\mathbb{R}^{K\times D\times C}\) on the shared event grid
\(\{t_k\}_{k=1}^{K}\) (Sec.~\ref{sec:SED-SE}). To control computation and memory while retaining event salience (the presence of spikes that mark informative observations), we downsample along the event axis via windowed max pooling, which preserves whether any spike occurs within each window and thus compresses long gaps. Let the pooling stride be \(s\in\mathbb{N}\) and define \(K'=\lfloor K/s\rfloor\).
For pooled index \(u\in\{1,\dots,K'\}\), variate \(d\in\{1,\dots,D\}\), and channel
\(c\in\{1,\dots,C\}\), set:
\begin{equation}
\label{eq:pool-spike}
\mathbf{S}'[u,d,c]
\;=\;
\max_{k\in\{(u-1)s+1,\dots,us\}}\;\mathbf{S}[k,d,c],\;\;
\mathbf{S}'\in\mathbb{R}^{K'\times D\times C}.
\end{equation}
We align pooled tokens to time by assigning the window’s last event time
\(t'_u=t_{us}\), producing a pooled grid \(\{t'_u\}_{u=1}^{K'}\). If an observation
mask \(\mathbf{M}\in\{0,1\}^{K\times D}\) accompanies the inputs, we downsample it
with the same rule
\(
\mathbf{M}'[u,d]=\max_{k\in\{(u-1)s+1,\dots,us\}}\mathbf{M}[k,d]
\)
so that a pooled position is marked observed if any event in its window is observed.
This reduces the effective event length from \(K\) to \(K'\) and yields a near-linear
reduction in compute and memory for the backbone while preserving within-window
spike occurrences.

To keep timestamps aligned with the pooled tokens, we assign the window’s last event time to the pooled position \(t'_{u}\;=\;t_{us}, \{t'_{u}\}_{u=1}^{K'}.\)
If an observation mask \(\mathbf{M}\in\{0,1\}^{K\times D}\) (1 if observed, 0
otherwise) accompanies the inputs, we downsample it by window-wise max so that a
pooled token is marked observed if any event in its window is observed:
\begin{equation}
\label{eq:pool-mask}
\mathbf{M}'[u,d]
\;=\;
\max_{k\in\{(u-1)s+1,\dots,us\}}\;\mathbf{M}[k,d],\;
\mathbf{M}'\in\{0,1\}^{K'\times D}.
\end{equation}
This procedure reduces the effective event length from \(K\) to
\(K'=\lfloor K/s\rfloor\), yielding a near-linear reduction in compute and memory
for the backbone while preserving within-window event salience. The pooled
spikes \(\mathbf{S}'\) and aligned timestamps \(\{t'_u\}\) form the input sequence for the subsequent SNN backbone in
Sec.~\ref{sec:sed-st}, which operates directly on this event-preserving,
downsampled representation.

\subsection{SED-based Spike Transformer Backbone}
\label{sec:sed-st}

After event-preserving downsampling (Sec.~\ref{sec:downsample}), we obtain spikes
\(\mathbf{S}'\in\mathbb{R}^{K'\times D\times C}\) to represent the IMTS on the pooled event grid
\(\mathcal{T}'=\{t'_u\}_{u=1}^{K'}\). Subsequently, we design the temporal spiking neural network, namely SED-based Spike Transformer (SED-ST), as the backbone to model the obtained spike trains. Technically, SED-ST learns intra-series temporal representations in an
event-synchronous manner: at each pooled event step \(u\) we treat the \(D\) variates as \(D\) tokens,
embed their spike channels, enrich them with time, and pass the resulting \(D\times d\) token matrix
through \(L\) identical blocks. This design preserves event semantics by updating only at \(\mathcal{T}'\), and exploits sparsity with a linear-time attention whose
scores are explicitly conditioned on inter-event gaps. For clarity, we omit the batch dimension $B$ in what follows. Let \(E\in\mathbb{R}^{C\times d}\) be a learnable linear projection that maps the \(C\) spike channels to the model width \(d\) (shared across variates and time), and let \(\mathrm{TE}\) be a learnable time embedding. The token for variate \(d\) at pooled event step \(u\) is: 
\begin{equation}
\label{eq:sedst-embed}
\mathbf{x}_d[u] = E\,\mathbf{S}'[u,d,:] + \mathrm{TE}(t'_u) \in \mathbb{R}^{d},
\qquad
E\in\mathbb{R}^{C\times d},
\end{equation}
stacking across variates yields
\(\mathbf{X}[u]\!=\![\mathbf{x}_1[u],\dots,\mathbf{x}_D[u]]^\top\!\in\!\mathbb{R}^{D\times d}\), where $\mathbf{X}[u]$ are continuous values rather than discrete spike trains.

\subsubsection{SED-Attention: membrane-based linear attention.}
To achieve efficient computation while remaining event-aware, we combine per-head projections with
EA-LIF feature maps and propose the SED-Attention (SED-A) mechanism. With \(H\) heads and \(d_h=d/H\), we apply normalized projections:
\[
\mathbf{q}[u]\!=\!\mathrm{BN}_q(W_q\mathbf{X}[u]),\
\mathbf{k}[u]\!=\!\mathrm{BN}_k(W_k\mathbf{X}[u]),\
\mathbf{v}[u]\!=\!\mathrm{BN}_v(W_v\mathbf{X}[u]),
\]
and split row-wise into heads
\(\mathbf{q}_h[u],\mathbf{k}_h[u],\mathbf{v}_h[u]\in\mathbb{R}^{D\times d_h}\).
We then obtain non-negative, event-aware query and key via the EA-LIF mapping
(Sec.~\ref{sec:ea-lif}) on the pooled grid, with \(\Delta t'_u=t'_u-t'_{u-1}\ge 0\):
\begin{equation}
\boldsymbol{\phi}_h^{(q)}[u]\!=\!\mathcal{EA\mbox{-}LIF}^{\mathrm{(w/\,softplus)}}\big(\mathbf{q}_h[u];\Delta t'_u\big),
\end{equation}
\begin{equation}
\boldsymbol{\phi}_h^{(k)}[u] \!=\! \mathcal{EA\mbox{-}LIF}^{\mathrm{(w/\,softplus)}}\big(\mathbf{k}_h[u];\Delta t'_u\big).
\end{equation}
For the value stream we keep the same interval-conditioned leak but omit the final
non-negativity squash:
\begin{equation}
\label{eq:sedst-ealif-v}
\tilde{\mathbf{v}}_h[u]\!=\! \mathcal{EA\mbox{-}LIF}^{\mathrm{(w/o\,softplus)}}\!\big(\mathbf{v}_h[u];\,\Delta t'_u\big)
\in\mathbb{R}^{D\times d_h}.
\end{equation}
Using the standard linear-attention kernel trick \cite{katharopoulos2020transformers,comba}, the head-wise output at step \(u\) is:
\begin{equation}
\label{eq:sedst-linatt}
\mathbf{y}_h[u]
\;=\;
\frac{\;\boldsymbol{\phi}_h^{(q)}[u]\;\Big(\sum_{w=1}^{K'} \boldsymbol{\phi}_h^{(k)}[w]^\top \tilde{\mathbf{v}}_h[w]\Big)\;}
     {\;\big(\boldsymbol{\phi}_h^{(q)}[u]\;\sum_{w=1}^{K'} \boldsymbol{\phi}_h^{(k)}[w]^\top\big)\;}
\;\in\;\mathbb{R}^{D\times d_h},
\end{equation}
where the pre-aggregates
\(\sum_{w}\boldsymbol{\phi}_h^{(k)}[w]\) and
\(\sum_{w}\boldsymbol{\phi}_h^{(k)}[w]^\top\tilde{\mathbf{v}}_h[w]\)
are computed once per block. Concatenating heads and projecting:
\begin{equation}
\label{eq:sedst-head-merge}
\mathbf{Y}[u] \;=\; W_o\,\cdot\mathrm{Concat}_h\big(\mathbf{y}_h[u]\big)\;\in\;\mathbb{R}^{D\times d}.
\end{equation}
Overall complexity is linear in $K'\!:\!\mathcal{O}(K'H{d_h}^2)$ where attention scores depend
explicitly on \(\Delta t'_u\). We drop the softmax used in vanilla self-attention, realizing both SED and linear complexity within the attention mechanism.

\subsubsection{Block ordering and masked time aggregation.}
Each SED-ST block adopts a pre-norm Membrane Shortcut~\cite{MembraneShortcut,fang2025spikingtransformersneedhigh}
with two sequential updates. First, features computed by SED-A on a batch-normalized input are added
back via a shortcut:
\begin{equation}
\label{eq:sedst-block-attn}
\widetilde{\mathbf{X}}^{(\ell)}[u]
\!=\!
\mathbf{X}^{(\ell)}[u]
\!+\!
\mathrm{SED\mbox{-}A}\big(\mathrm{BN}(\mathbf{X}^{(\ell)}[u]);\mathcal{T}'\big),
\ell=0,\dots,L\!-\!1.
\end{equation}
Second, a position-wise feed-forward network (FFN) is applied and added back via a shortcut to form
the block output:
\begin{equation}
\label{eq:sedst-block-ffn}
\mathbf{X}^{(\ell+1)}[u]
\;=\;
\widetilde{\mathbf{X}}^{(\ell)}[u]
\;+\;
\mathrm{FFN}\big(\mathrm{BN}(\widetilde{\mathbf{X}}^{(\ell)}[u])\big).
\end{equation}

After stacking \(L\) blocks, we perform Masked Time Aggregation (MTA) to summarize each variate over
the pooled event axis while avoiding bias from unobserved steps. Using \(\mathbf{M}'\in\{0,1\}^{K'\times D}\):
\begin{equation}
\label{eq:sedst-agg}
\mathbf{z}[d] \;=\;
\frac{\sum_{u=1}^{K'} \mathbf{M}'[u,d]\;\mathbf{X}^{(L)}[u,d,:]}
     {\sum_{u=1}^{K'} \mathbf{M}'[u,d]}
\in\mathbb{R}^{d},
\;\; d=1,\dots,D.
\end{equation}
This aggregation uses only observed events (entries with \(\mathbf{M}'[u,d]=1\)), respecting irregular sampling and preventing missing windows from diluting the representation. Therefore, MTA produces a length-agnostic, missingness-robust summary aligned to the event grid, providing a compact input for downstream prediction.

\begin{table*}[!ht]
\centering
\small
\renewcommand{\arraystretch}{1} 
\setlength{\tabcolsep}{4pt}
\caption{
Forecasting results (MSE$\downarrow$, MAE$\downarrow$) of baselines and SEDformer on the sparsified Wiki2000 and WikiArticle datasets under different Sparsifying Rates. All methods use 90 historical days to predict the next 30 days. 
Scores are reported as mean~$\pm$~std over repeated runs with 5 different seeds. 
The best result in each column is shown in \textbf{bold}, and the second best is \underline{underlined}. All numbers come from our unified re-implementation using each method’s recommended settings for fair comparison.}
\vspace{-3mm}
\label{tab:results90-30}
\scalebox{0.82}{
\begin{tabular}{c|cc|cc|cc|cc|cc|cc}
\toprule
\multirow{3}{*}{\textbf{Model}} 
& \multicolumn{6}{c|}{\textbf{Wiki2000}} 
& \multicolumn{6}{c}{\textbf{WikiArticle}} \\
\cmidrule(lr){2-7}\cmidrule(lr){8-13}
& \multicolumn{2}{c|}{\textbf{Sparsifying Rate = 25\%}} 
& \multicolumn{2}{c|}{\textbf{Sparsifying Rate = 50\%}} 
& \multicolumn{2}{c|}{\textbf{Sparsifying Rate = 75\%}}
& \multicolumn{2}{c|}{\textbf{Sparsifying Rate = 25\%}} 
& \multicolumn{2}{c|}{\textbf{Sparsifying Rate = 50\%}} 
& \multicolumn{2}{c}{\textbf{Sparsifying Rate = 75\%}} \\
 \cmidrule(lr){1-1}\cmidrule(lr){2-3}\cmidrule(lr){4-5}\cmidrule(lr){6-7}\cmidrule(lr){8-9}\cmidrule(lr){10-11}\cmidrule(lr){12-13}
 \textbf{Metric} & \textbf{MSE$\times 10^{-3}$} & \textbf{MAE$\times 10^{-2}$} & \textbf{MSE$\times 10^{-2}$} & \textbf{MAE$\times 10^{-2}$} & \textbf{MSE$\times 10^{-2}$} & \textbf{MAE$\times 10^{-2}$}
& \textbf{MSE$\times 10^{-3}$} & \textbf{MAE$\times 10^{-2}$} & \textbf{MSE$\times 10^{-2}$} & \textbf{MAE$\times 10^{-2}$} & \textbf{MSE$\times 10^{-2}$} & \textbf{MAE$\times 10^{-2}$} \\
\midrule
Neural ODE & 13.50 {\scriptsize $\pm$ 0.02} & 8.06 {\scriptsize $\pm$ 0.12} & 2.27 {\scriptsize $\pm$ 0.06} & 11.10 {\scriptsize $\pm$ 0.12} & 3.33 {\scriptsize $\pm$ 0.07} & 13.50 {\scriptsize $\pm$ 0.02} & 9.52 {\scriptsize $\pm$ 0.12} & 6.43 {\scriptsize $\pm$ 0.01} & 2.13 {\scriptsize $\pm$ 0.07} & 9.26 {\scriptsize $\pm$ 0.08} & 3.63 {\scriptsize $\pm$ 0.06} & 11.33 {\scriptsize $\pm$ 0.04} \\
Latent ODE & 13.40 {\scriptsize $\pm$ 0.04} & 7.76 {\scriptsize $\pm$ 0.06} & 1.58 {\scriptsize $\pm$ 0.04} & 9.52 {\scriptsize $\pm$ 0.02} & 2.40 {\scriptsize $\pm$ 0.06} & 12.02 {\scriptsize $\pm$ 0.12} & 9.93 {\scriptsize $\pm$ 0.06} & 7.46 {\scriptsize $\pm$ 0.10} & 1.39 {\scriptsize $\pm$ 0.05} & 7.46 {\scriptsize $\pm$ 0.04} & 2.09 {\scriptsize $\pm$ 0.02} & 9.41 {\scriptsize $\pm$ 0.05} \\
GRU-ODE & 10.28 {\scriptsize $\pm$ 0.03} & 7.21 {\scriptsize $\pm$ 0.11} & 1.69 {\scriptsize $\pm$ 0.04} & 9.35 {\scriptsize $\pm$ 0.13} & 2.35 {\scriptsize $\pm$ 0.08} & 11.62 {\scriptsize $\pm$ 0.03} & 9.38 {\scriptsize $\pm$ 0.11} & 6.58 {\scriptsize $\pm$ 0.02} & 2.08 {\scriptsize $\pm$ 0.06} & 9.41 {\scriptsize $\pm$ 0.09} & 2.65 {\scriptsize $\pm$ 0.07} & 11.28 {\scriptsize $\pm$ 0.05} \\
GRU-D   & 10.55 {\scriptsize $\pm$ 0.02} & 8.12 {\scriptsize $\pm$ 0.15} & 2.41 {\scriptsize $\pm$ 0.08} & 9.95 {\scriptsize $\pm$ 0.10} & 3.48 {\scriptsize $\pm$ 0.09} & 9.91 {\scriptsize $\pm$ 0.04} & 9.67 {\scriptsize $\pm$ 0.13} & 6.48 {\scriptsize $\pm$ 0.03} & 2.05 {\scriptsize $\pm$ 0.05} & 9.73 {\scriptsize $\pm$ 0.11} & 2.48 {\scriptsize $\pm$ 0.05} & 9.65 {\scriptsize $\pm$ 0.06} \\
SeFT   & 10.23 {\scriptsize $\pm$ 0.03} & 6.52 {\scriptsize $\pm$ 0.10} & 2.08 {\scriptsize $\pm$ 0.04} & 11.88 {\scriptsize $\pm$ 0.14} & 3.12 {\scriptsize $\pm$ 0.06} & 12.76 {\scriptsize $\pm$ 0.05} & 8.95 {\scriptsize $\pm$ 0.09} & 7.01 {\scriptsize $\pm$ 0.04} & 2.35 {\scriptsize $\pm$ 0.09} & 8.84 {\scriptsize $\pm$ 0.07} & 1.91 {\scriptsize $\pm$ 0.08} & 8.98 {\scriptsize $\pm$ 0.03} \\
RainDrop   & 13.72 {\scriptsize $\pm$ 0.05} & 8.38 {\scriptsize $\pm$ 0.13} & 1.89 {\scriptsize $\pm$ 0.03} & 10.73 {\scriptsize $\pm$ 0.09} & 3.47 {\scriptsize $\pm$ 0.10} & 14.15 {\scriptsize $\pm$ 0.06} & 9.21 {\scriptsize $\pm$ 0.14} & 6.96 {\scriptsize $\pm$ 0.02} & 2.28 {\scriptsize $\pm$ 0.08} & 9.58 {\scriptsize $\pm$ 0.10} & 3.29 {\scriptsize $\pm$ 0.04} & 11.87 {\scriptsize $\pm$ 0.07} \\
Warpformer  & 13.04 {\scriptsize $\pm$ 0.02} & 8.67 {\scriptsize $\pm$ 0.11} & 2.12 {\scriptsize $\pm$ 0.07} & 9.45 {\scriptsize $\pm$ 0.11} & 2.98 {\scriptsize $\pm$ 0.05} & 10.28 {\scriptsize $\pm$ 0.03} & 9.16 {\scriptsize $\pm$ 0.10} & 6.75 {\scriptsize $\pm$ 0.03} & 1.97 {\scriptsize $\pm$ 0.04} & 9.02 {\scriptsize $\pm$ 0.06} & 2.81 {\scriptsize $\pm$ 0.09} & 9.99 {\scriptsize $\pm$ 0.02} \\
mTAND & 9.98 {\scriptsize $\pm$ 0.03} & 9.22 {\scriptsize $\pm$ 0.09} & 1.95 {\scriptsize $\pm$ 0.06} & 8.31 {\scriptsize $\pm$ 0.03} & 2.89 {\scriptsize $\pm$ 0.11} & 9.45 {\scriptsize $\pm$ 0.12} & 10.34 {\scriptsize $\pm$ 0.05} & 8.12 {\scriptsize $\pm$ 0.11} & 2.03 {\scriptsize $\pm$ 0.03} & 7.45 {\scriptsize $\pm$ 0.05} & 2.18 {\scriptsize $\pm$ 0.04} & 9.01 {\scriptsize $\pm$ 0.05} \\
CRU & 11.25 {\scriptsize $\pm$ 0.02} & 9.40 {\scriptsize $\pm$ 0.08} & 1.80 {\scriptsize $\pm$ 0.05} & 8.03 {\scriptsize $\pm$ 0.02} & 2.68 {\scriptsize $\pm$ 0.10} & 9.10 {\scriptsize $\pm$ 0.11} & 10.02 {\scriptsize $\pm$ 0.04} & 7.78 {\scriptsize $\pm$ 0.10} & 1.86 {\scriptsize $\pm$ 0.02} & 7.12 {\scriptsize $\pm$ 0.04} & 2.01 {\scriptsize $\pm$ 0.03} & 8.68 {\scriptsize $\pm$ 0.04} \\
Neural Flow & 10.35 {\scriptsize $\pm$ 0.03} & 8.32 {\scriptsize $\pm$ 0.09} & 1.76 {\scriptsize $\pm$ 0.06} & 7.12 {\scriptsize $\pm$ 0.12} & 2.20 {\scriptsize $\pm$ 0.08} & 8.90 {\scriptsize $\pm$ 0.16} & 10.12 {\scriptsize $\pm$ 0.05} & 7.12 {\scriptsize $\pm$ 0.08} & 1.66 {\scriptsize $\pm$ 0.02} & 6.90 {\scriptsize $\pm$ 0.05} & 1.91 {\scriptsize $\pm$ 0.04} & 8.18 {\scriptsize $\pm$ 0.06} \\
tPatchGNN & 9.88 {\scriptsize $\pm$ 0.03} & 5.89 {\scriptsize $\pm$ 0.11} & 1.49 {\scriptsize $\pm$ 0.12} & 6.68 {\scriptsize $\pm$ 0.12} & \underline{1.85 {\scriptsize $\pm$ 0.10}} & \underline{8.60 {\scriptsize $\pm$ 0.12}} & 8.81 {\scriptsize $\pm$ 0.10} & 5.38 {\scriptsize $\pm$ 0.08} & \underline{1.18 {\scriptsize $\pm$ 0.03}} & \underline{6.10 {\scriptsize $\pm$ 0.06}} & 1.83 {\scriptsize $\pm$ 0.05} & 7.89 {\scriptsize $\pm$ 0.05} \\
GraFITi & 9.65 {\scriptsize $\pm$ 0.08} & 6.20 {\scriptsize $\pm$ 0.06} & 1.60 {\scriptsize $\pm$ 0.02} & 7.02 {\scriptsize $\pm$ 0.10} & 1.90 {\scriptsize $\pm$ 0.06} & 8.81 {\scriptsize $\pm$ 0.06} & 9.62 {\scriptsize $\pm$ 0.04} & 6.02 {\scriptsize $\pm$ 0.06} & 1.46 {\scriptsize $\pm$ 0.04} & 6.61 {\scriptsize $\pm$ 0.03} & 1.90 {\scriptsize $\pm$ 0.03} & 8.12 {\scriptsize $\pm$ 0.08} \\
TimeCHEAT & 10.35 {\scriptsize $\pm$ 0.12} & 6.01 {\scriptsize $\pm$ 0.08} & 1.72 {\scriptsize $\pm$ 0.04} & 7.08 {\scriptsize $\pm$ 0.03} & 2.02 {\scriptsize $\pm$ 0.04} & 9.80 {\scriptsize $\pm$ 0.10} & 10.02 {\scriptsize $\pm$ 0.08} & 6.12 {\scriptsize $\pm$ 0.04} & 1.60 {\scriptsize $\pm$ 0.06} & 7.12 {\scriptsize $\pm$ 0.06} & 2.20 {\scriptsize $\pm$ 0.03} & 8.03 {\scriptsize $\pm$ 0.06} \\
KAFNet & 9.79 {\scriptsize $\pm$ 0.19} & 5.75 {\scriptsize $\pm$ 0.12} & \underline{1.44 {\scriptsize $\pm$ 0.03}} & 6.78 {\scriptsize $\pm$ 0.23} & 1.86 {\scriptsize $\pm$ 0.04} & 8.66 {\scriptsize $\pm$ 0.23} & 9.10 {\scriptsize $\pm$ 0.16} & 5.37 {\scriptsize $\pm$ 0.08} & 1.19 {\scriptsize $\pm$ 0.02} & 6.23 {\scriptsize $\pm$ 0.06} & 1.88 {\scriptsize $\pm$ 0.02} & 7.92 {\scriptsize $\pm$ 0.07} \\
HyperIMTS & \underline{9.42 {\scriptsize $\pm$ 0.03 }} & \underline{ 5.40 {\scriptsize $\pm$ 0.10 }} & 1.56{\scriptsize $\pm$ 0.22 } & \underline{ 6.64{\scriptsize $\pm$ 0.07 }}&  1.90{\scriptsize $\pm$ 0.04 } & 8.70{\scriptsize $\pm$ 0.03 } & \underline{8.80{\scriptsize $\pm$ 0.06}} & \underline{5.30{\scriptsize $\pm$ 0.08}} & 1.25{\scriptsize $\pm$ 0.09 } & 6.15{\scriptsize $\pm$ 0.08 } & \underline{1.78{\scriptsize $\pm$ 0.01}} & \underline{7.88{\scriptsize $\pm$ 0.08}}  \\
\textbf{SEDformer} & \textbf{ 9.19 {\scriptsize $\pm$ 0.13 }} & \textbf{ 5.24 {\scriptsize $\pm$ 0.11 }} & \textbf{ 1.36{\scriptsize $\pm$ 0.22 }} & \textbf{ 6.44{\scriptsize $\pm$ 0.17 }}& \textbf{ 1.81{\scriptsize $\pm$ 0.02 }} & \textbf{ 8.53{\scriptsize $\pm$ 0.02 }} & \textbf{ 8.71{\scriptsize $\pm$ 0.26 }} & \textbf{ 5.21{\scriptsize $\pm$ 0.09 }} & \textbf{ 1.15{\scriptsize $\pm$ 0.18 }} & \textbf{ 6.00{\scriptsize $\pm$ 0.18 }} & \textbf{ 1.69{\scriptsize $\pm$ 0.02 }} & \textbf{ 7.84{\scriptsize $\pm$ 0.05 }}  \\
\bottomrule
\end{tabular}}
\vspace{-2mm}
\end{table*}

\begin{table*}[!ht]
\centering
\small
\renewcommand{\arraystretch}{0.9} 
\setlength{\tabcolsep}{4pt}
\caption{
Statistical significance test results (paired t-test) comparing SEDformer against the second-best method HyperIMTS. P-values less than 0.05 indicate statistically significant improvements.}
\vspace{-3mm}
\label{tab:Statisticaltest}
\scalebox{0.86}{
\begin{tabular}{c|cc|cc|cc|cc|cc|cc}
\toprule
\multirow{3}{*}{\textbf{Dataset}} 
& \multicolumn{6}{c|}{\textbf{Wiki2000}} 
& \multicolumn{6}{c}{\textbf{WikiArticle}} \\
\cmidrule(lr){2-7}\cmidrule(lr){8-13}
& \multicolumn{2}{c|}{\textbf{Sparsifying Rate = 25\%}} 
& \multicolumn{2}{c|}{\textbf{Sparsifying Rate = 50\%}} 
& \multicolumn{2}{c|}{\textbf{Sparsifying Rate = 75\%}}
& \multicolumn{2}{c|}{\textbf{Sparsifying Rate = 25\%}} 
& \multicolumn{2}{c|}{\textbf{Sparsifying Rate = 50\%}} 
& \multicolumn{2}{c}{\textbf{Sparsifying Rate = 75\%}} \\
 \cmidrule(lr){1-1}\cmidrule(lr){2-3}\cmidrule(lr){4-5}\cmidrule(lr){6-7}\cmidrule(lr){8-9}\cmidrule(lr){10-11}\cmidrule(lr){12-13}
 \textbf{Metric} & \textbf{MSE} & \textbf{MAE} & \textbf{MSE} & \textbf{MAE} & \textbf{MSE} & \textbf{MAE}
& \textbf{MSE} & \textbf{MAE} & \textbf{MSE} & \textbf{MAE} & \textbf{MSE} & \textbf{MAE} \\
\midrule
\textbf{P-value} & 0.038 & 0.021 & 0.015 & 0.032 & 0.008 & 0.003 & 0.042 & 0.028 & 0.012 & 0.019 & 0.002 & 0.045 \\
\bottomrule
\end{tabular}}
\vspace{-3mm}
\end{table*}

\subsection{Query\mbox{-}Conditioned Decoder}
\label{sec:decoder}

\subsubsection{Answering the Queries}
After event irregularity has been compressed by the EPTD module and intra\mbox{-}series dependencies have been refined by the $L$ stacked SED\mbox{-}ST blocks, our model outputs the representation $\mathbf{Z}\in\mathbb{R}^{D\times d}$; its $j$-th row $\mathbf{z}_{j}$ is a compact summary of variate $j$. For each query time $q^{(r)}_{j}$, we concatenate this summary with its learnable time embedding and map the result to a scalar prediction via an MLP:
\begin{equation}
\hat{x}^{\,j}_{r}\;=\;\mathrm{MLP}\Bigl(\mathbf{z}_{j}\oplus\mathrm{TE}(q^{(r)}_{j})\Bigr),\;\;\; j=1,\ldots,D,\;\; r=1,\ldots,Q_j.
  \label{eq:forecast}
\end{equation}
The head is query\mbox{-}conditioned (each output depends on its own query time) and parallelizable across all $(j,r)$, while remaining lightweight since all cross\mbox{-}time interactions are handled upstream.

\subsubsection{Training objective.}
Given ground\mbox{-}truth values $\{x^{\,j}_{r}\}$ at the query times $\{q^{(r)}_{j}\}$, we minimize mean squared error (MSE) averaged over variates and their queries:
\begin{equation}
\mathcal{L}
\;=\;
\frac{1}{D}\sum_{j=1}^{D}\;\frac{1}{Q_j}\sum_{r=1}^{Q_j}
\Big(\,\hat{x}^{\,j}_{r}-x^{\,j}_{r}\,\Big)^{2}.
\label{eq:loss}
\end{equation}

\section{Experiments}

\subsection{Experimental Setup}

\subsubsection{Datasets.} 
We experiment on two widely used public telemetry IMTS benchmarks (originating from web-traffic telemetry) that exhibit long missing spans and bursty event patterns common in Internet-connected telemetry streams. 
\textbf{Wiki2000} \cite{Wike2000}: daily page--view counts for 9{,}013 Wikipedia pages. We use the TFB release \cite{qiu2024tfb} and select the first 200 variates. 
\textbf{WikiArticle} \cite{WikeArticle}: the Kaggle \emph{Web Traffic Time Series Forecasting} set; we load \texttt{train\_1.csv} and keep the first 300 variates. 
We restrict to 200/300 series to ensure parity of training cost across baselines and reproducibility; choosing the ``first'' in the provider's order avoids cherry-picking and yields a fixed, bias-free subset. 
Both raw corpora contain long missing spans and sporadic anomalies that make direct forecasting ill-posed. Following common benchmarking practice, we first create a clean reference trajectory per variate on a regular daily grid and then introduce irregularity in a controlled way \cite{SUSTeR}. 
Concretely, we perform local gap filling via second-order Lagrange interpolation and mild outlier smoothing, and subsequently ``sparsify'' the cleaned series by independently masking observations at rates 25\%/50\%/75\%. 
This simulation of data sparsity via independent, random masking corresponds to the Missing Completely at Random (MCAR) mechanism, a standard and foundational approach in benchmarking literature for creating controlled, unbiased evaluation scenarios \cite{jager2021benchmark}. 
This two-step protocol has been used to evaluate IMTS models under controlled irregular sampling while keeping the latent signal fixed for fair comparison \cite{SUSTeR,du2024tsibench}. 
It also aligns with continuous-time evaluations that simulate non-uniform sampling or random deletions \cite{Latent-ODE,gpvae2020}. 
We refer to the masking rate as the ``Sparsifying Rate'', and introduce the detailed procedure of constructing these datasets in Appendix~\ref{app:datasets}. 
We adopt a multi-step setting: a 90-day history window is used to forecast the next 30 days.

\begin{figure*}[t]
  \captionsetup[sub]{skip=0.2pt}  
  \centering
  {\includegraphics[width=0.23\linewidth]{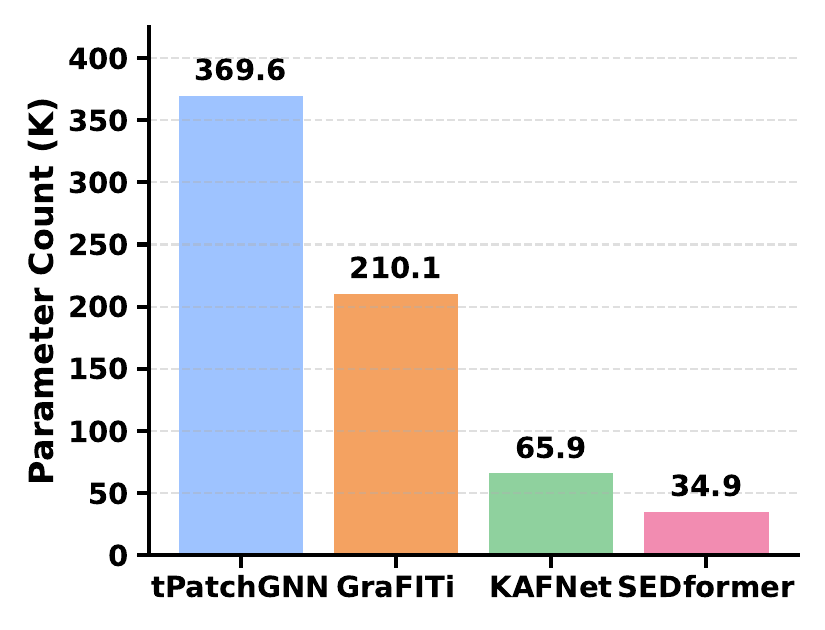}%
   \label{fig:num_parameters}}
  {\includegraphics[width=0.23\linewidth]{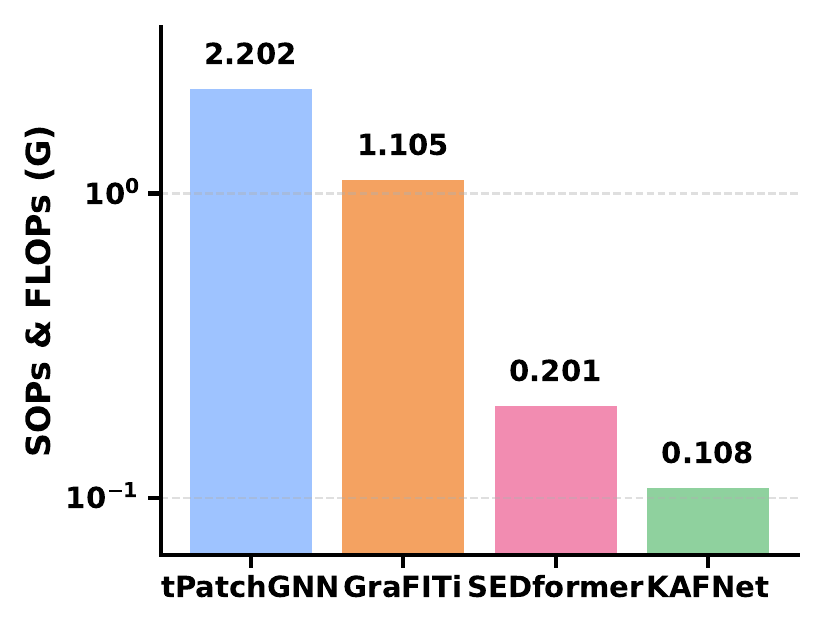}%
   \label{fig:flops}}
  {\includegraphics[width=0.23\linewidth]{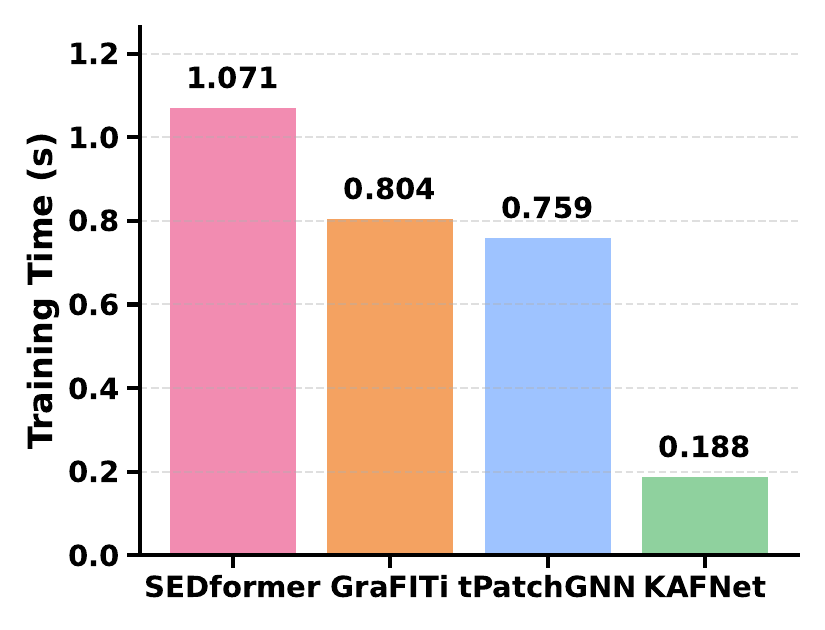}%
   \label{fig:training_time}}
  {\includegraphics[width=0.23\linewidth]{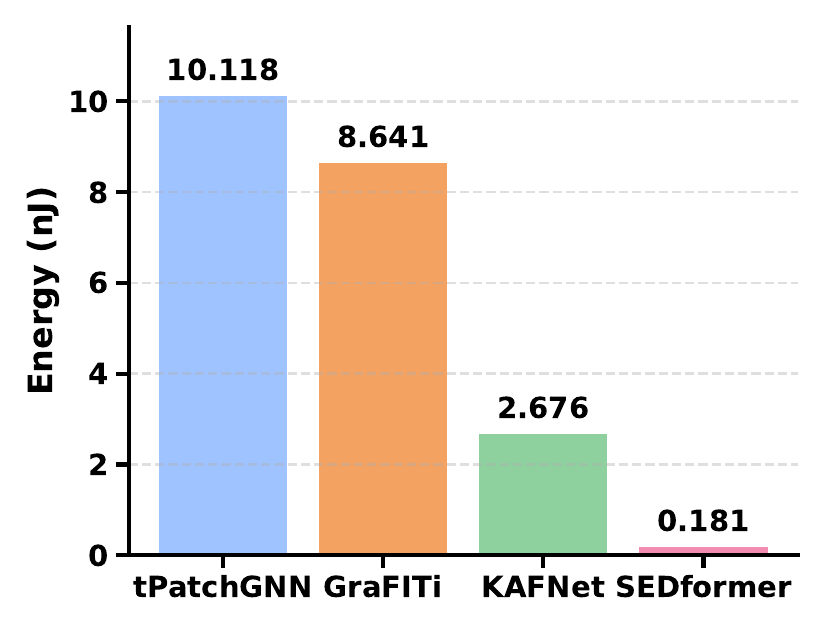}
   \label{fig:inference_time}}
  \vspace{-4mm}
  \caption{Efficiency analysis on the Wiki2000 dataset with Sparsifying Rate = $\mathbf{25\%}$. “OPs” refers to SOPs in SNN and FLOPs in ANN. “SOPs” is the synaptic operations of SEDformer. “FLOPs” denotes the floating point operations of other ANN baselines.}
  \vspace{-4mm}
  \label{fig:efficiency}
\end{figure*}

\vspace{-2mm}

\subsubsection{Baselines.}
Telemetry streams are naturally sparse and irregularly sampled, with large missing observations and non-uniform intervals. \textbf{Unlike prior works that treat them as dense continuous streams \cite{ContiFormer}, the IMTS formulation explicitly models discrete events with irregular timestamps and better captures their true structure.} Following the standard IMTS forecasting setting (Sec.~\ref{sec:relatedworks}), we treat telemetry forecasting as an \textbf{offline sequence-to-sequence} problem (Sec.~\ref{subsec:problem_formulation}) and evaluate SEDformer against strong IMTS baselines from four methodological classes. (i) RNN-based: GRU-D \cite{Che2018}. (ii) Differential Equation–based: Neural ODE \cite{Neural-ODE}, Latent ODE \cite{Latent-ODE}, GRU-ODE \cite{GRU-ODE}, CRU \cite{CRU}, and Neural Flow \cite{Neural-Flows}. (iii) Transformer– and Attention–based: ContiFormer \cite{ContiFormer}, Warpformer \cite{Warpformer}, mTAND \cite{mTAND}, SeFT \cite{seft}, and KAFNet \cite{kafnet}. (iv) GNN– and Set-based: RainDrop \cite{RainDrop}, GraFITi \cite{GraFITi}, tPatchGNN \cite{tPatchGNN}, TimeCHEAT \cite{TimeCHEAT}, and HyperIMTS \cite{HyperIMTS}. For methods originally proposed for IMTS interpolation (e.g., mTAND \cite{mTAND}), we swap interpolation targets for queries to enable extrapolation. For methods originally proposed for IMTS classification (e.g., SeFT \cite{seft}, Warpformer \cite{Warpformer}), we replace the classification head with a MLP forecasting head so that all baselines output real–valued trajectories under the same training objective. A detailed introduction is provided in Appendix \ref{app:baselines}.

 \vspace{-2mm}

\subsubsection{Implementation Details.}\label{sec:implement}
We run all experiments on a single NVIDIA RTX~3090 GPU.  
Training uses the Adam optimizer \cite{kingma2017adam}. Model quality is reported with two standard metrics, MSE and MAE, computed over the set of queried timestamps $\mathcal{Q}$ as \(\text{MSE}=\frac{1}{|\mathcal{Q}|}\sum_{i=1}^{|\mathcal{Q}|}\!\bigl(x_i-\hat{x}_i\bigr)^2,\;\text{MAE}=\frac{1}{|\mathcal{Q}|}\sum_{i=1}^{|\mathcal{Q}|}\!\bigl|x_i-\hat{x}_i\bigr|.\) Here $x_i$ and $\hat{x}_i$ denote the ground truth and the prediction at the $i$-th query, respectively, and $|\mathcal{Q}|$ is the number of queries. The complete hyperparameter grids and selection protocol are provided in Appendix \ref{app:hyperparameter}.

\vspace{-2mm}

\subsection{Main Results}

\label{sec:main_results}
\noindent
We summarize the full comparisons in Tab.~\ref{tab:results90-30}. Overall, SEDformer attains the strongest accuracy across both datasets under a wide range of Sparsifying Rates, with the gains most pronounced in the medium–to–high sparsity regimes that best reflect real telemetry streams. We attribute this advantage to three design aspects: (i) Sparsity Awareness: the model naturally consumes irregular observations with event-aware encoding rather than relying on heavy imputation, thus avoiding bias from ad hoc gap-filling; (ii) Multi-scale Temporal Modeling: local bursts and long-range seasonalities are jointly captured, which is crucial when observations are asynchronous and event-driven; and (iii) Cross-series Consolidation with Robustness: shared structure is leveraged without over-smoothing individual series, improving stability across seeds. In contrast, families tailored to regular grids (e.g., pure patching \cite{tPatchGNN} or warping Transformers \cite{Warpformer}) underutilize time gaps and degrade as sparsity increases; missingness-handling RNNs \cite{Che2018, CRU} encode gaps but struggle to maintain long-range dependencies and cross-series transfer; and graph-based baselines \cite{GraFITi} help under moderate sparsity but may assume denser contexts and thus plateau at higher sparsity. We also conduct repeated runs and two-sided significance tests comparing SEDformer against the strongest baseline per setting; the $p$-values in Tab.~\ref{tab:Statisticaltest} indicate statistically reliable improvements in multiple regimes and comparable performance elsewhere, confirming that SEDformer provides optimal forecasting under real telemetry IMTS while remaining stable in easier settings.

\vspace{-1mm}

\begin{figure}[!ht]
    \centering
    \includegraphics[width=0.9\linewidth, trim=22 23 20 15, clip]{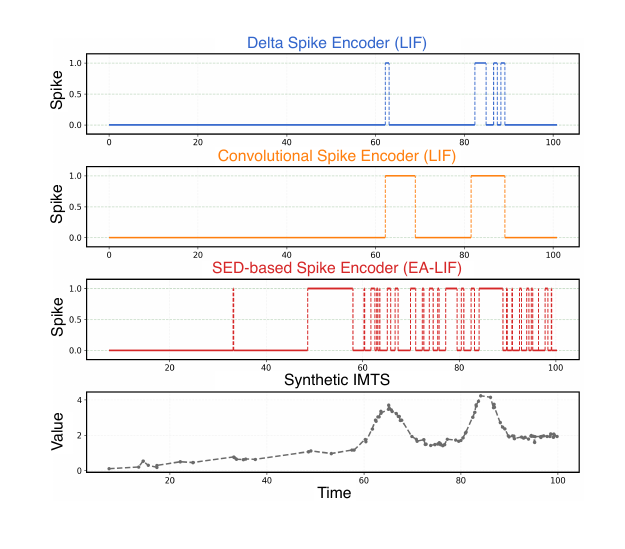}
    \vspace{-4mm}
    \caption{Spike trains generated by Delta Spike Encoder and Convolutional Spike Encoder with vanilla LIF neuron, as well as the proposed SED-based Spike Encoder with EA-LIF.}
    \label{fig:spiketrains}
    \vspace{-5mm}
\end{figure}

\vspace{-2mm}

\begin{table*}[t]
  \centering
  \caption{Ablation results of SEDformer on two datasets evaluated by MSE and MAE (mean±std). The best results are in \textbf{bold}.}
  \label{tab:ablation_results}
  \vspace{-12pt}
  \renewcommand{\arraystretch}{1.3} 
  \footnotesize
  \scalebox{0.78}{
  \begin{tabular}{c|cc|cc|cc|cc|cc|cc}
    \toprule
    \multirow{2}{*}{\textbf{Dataset}} 
    & \multicolumn{6}{c|}{\textbf{Wiki2000}} 
    & \multicolumn{6}{c}{\textbf{WikiArticle}} \\
    \cmidrule(lr){2-7}\cmidrule(lr){8-13}
    & \multicolumn{2}{c|}{\textbf{Sparsifying Rate = 25\%}} 
    & \multicolumn{2}{c|}{\textbf{Sparsifying Rate = 50\%}} 
    & \multicolumn{2}{c|}{\textbf{Sparsifying Rate = 75\%}}
    & \multicolumn{2}{c|}{\textbf{Sparsifying Rate = 25\%}} 
    & \multicolumn{2}{c|}{\textbf{Sparsifying Rate = 50\%}} 
    & \multicolumn{2}{c}{\textbf{Sparsifying Rate = 75\%}} \\
  \cmidrule(lr){1-1} \cmidrule(lr){2-3}\cmidrule(lr){4-5}\cmidrule(lr){6-7}\cmidrule(lr){8-9}\cmidrule(lr){10-11}\cmidrule(lr){12-13}
    \textbf{Metric} & \textbf{MSE$\times 10^{-3}$} & \textbf{MAE$\times 10^{-2}$} & \textbf{MSE$\times 10^{-2}$} & \textbf{MAE$\times 10^{-2}$} & \textbf{MSE$\times 10^{-2}$} & \textbf{MAE$\times 10^{-2}$}
    & \textbf{MSE$\times 10^{-3}$} & \textbf{MAE$\times 10^{-2}$} & \textbf{MSE$\times 10^{-2}$} & \textbf{MAE$\times 10^{-2}$} & \textbf{MSE$\times 10^{-2}$} & \textbf{MAE$\times 10^{-2}$} \\
    \midrule
    \textbf{SEDformer} & \textbf{ 9.19 {\scriptsize $\pm$ 0.13 }} & \textbf{ 5.24 {\scriptsize $\pm$ 0.11 }} & \textbf{ 1.36{\scriptsize $\pm$ 0.22 }} & \textbf{ 6.44{\scriptsize $\pm$ 0.17 }}& \textbf{ 1.81{\scriptsize $\pm$ 0.02 }} & \textbf{ 8.53{\scriptsize $\pm$ 0.02 }} & \textbf{ 8.71{\scriptsize $\pm$ 0.26 }} & \textbf{ 5.21{\scriptsize $\pm$ 0.09 }} & \textbf{ 1.15{\scriptsize $\pm$ 0.18 }} & \textbf{ 6.00{\scriptsize $\pm$ 0.18 }} & \textbf{ 1.69{\scriptsize $\pm$ 0.02 }} & \textbf{ 7.84{\scriptsize $\pm$ 0.05 }}  \\
    \midrule
    w/o SED-SE   & 10.91 {\scriptsize $\pm$ 0.03} & 5.51 {\scriptsize $\pm$ 0.07} & 2.60 {\scriptsize $\pm$ 0.03} & 6.91 {\scriptsize $\pm$ 0.06} & 1.96 {\scriptsize $\pm$ 0.10} & 8.92 {\scriptsize $\pm$ 0.08} & 10.08 {\scriptsize $\pm$ 0.04} & 6.56 {\scriptsize $\pm$ 0.08} & 1.47 {\scriptsize $\pm$ 0.03} & 6.20 {\scriptsize $\pm$ 0.05} & 2.26 {\scriptsize $\pm$ 0.05} & 8.52 {\scriptsize $\pm$ 0.05} \\
    w/o EA-LIF \& w/ LIF  & 10.95 {\scriptsize $\pm$ 0.03} & 6.03 {\scriptsize $\pm$ 0.05} & 1.70 {\scriptsize $\pm$ 0.04} & 7.78 {\scriptsize $\pm$ 0.02} & 2.55 {\scriptsize $\pm$ 0.10} & 8.85 {\scriptsize $\pm$ 0.09} & 9.78 {\scriptsize $\pm$ 0.05} & 7.62 {\scriptsize $\pm$ 0.10} & 1.78 {\scriptsize $\pm$ 0.03} & 6.89 {\scriptsize $\pm$ 0.05} & 1.92 {\scriptsize $\pm$ 0.04} & 8.52 {\scriptsize $\pm$ 0.02} \\
    w/o EPTD        &  10.98 {\scriptsize $\pm$ 0.02} & 6.09 {\scriptsize $\pm$ 0.04} & 1.74 {\scriptsize $\pm$ 0.05} & 7.88 {\scriptsize $\pm$ 0.03} & 2.61 {\scriptsize $\pm$ 0.11} & 8.96 {\scriptsize $\pm$ 0.12} & 9.81 {\scriptsize $\pm$ 0.04} & 7.63 {\scriptsize $\pm$ 0.11} & 1.80 {\scriptsize $\pm$ 0.02} & 6.99 {\scriptsize $\pm$ 0.04} & 1.95 {\scriptsize $\pm$ 0.03} & 8.55 {\scriptsize $\pm$ 0.05} \\     
    w/o SED-ST          & 11.08 {\scriptsize $\pm$ 0.02} & 6.18 {\scriptsize $\pm$ 0.04} & 1.76 {\scriptsize $\pm$ 0.05} & 7.94 {\scriptsize $\pm$ 0.03} & 2.63 {\scriptsize $\pm$ 0.11} & 9.01 {\scriptsize $\pm$ 0.12} & 10.29 {\scriptsize $\pm$ 0.03} & 7.68 {\scriptsize $\pm$ 0.11} & 1.82 {\scriptsize $\pm$ 0.02} & 7.03 {\scriptsize $\pm$ 0.04} & 2.27 {\scriptsize $\pm$ 0.04} & 8.76 {\scriptsize $\pm$ 0.05} \\
    w/o SED-A  \& w/ SA  & 10.81 {\scriptsize $\pm$ 0.03} & 6.25 {\scriptsize $\pm$ 0.08} & 1.69 {\scriptsize $\pm$ 0.04} & 7.75 {\scriptsize $\pm$ 0.02} & 2.54 {\scriptsize $\pm$ 0.10} & 8.83 {\scriptsize $\pm$ 0.11} & 9.65 {\scriptsize $\pm$ 0.05} & 7.49 {\scriptsize $\pm$ 0.10} & 1.75 {\scriptsize $\pm$ 0.03} & 6.86 {\scriptsize $\pm$ 0.05} & 1.96 {\scriptsize $\pm$ 0.05} & 8.42 {\scriptsize $\pm$ 0.04} \\
    \bottomrule
  \end{tabular}}
  \vspace{-2mm}
\end{table*}

\begin{figure*}[t]
  \centering
  {\includegraphics[width=0.23\linewidth]{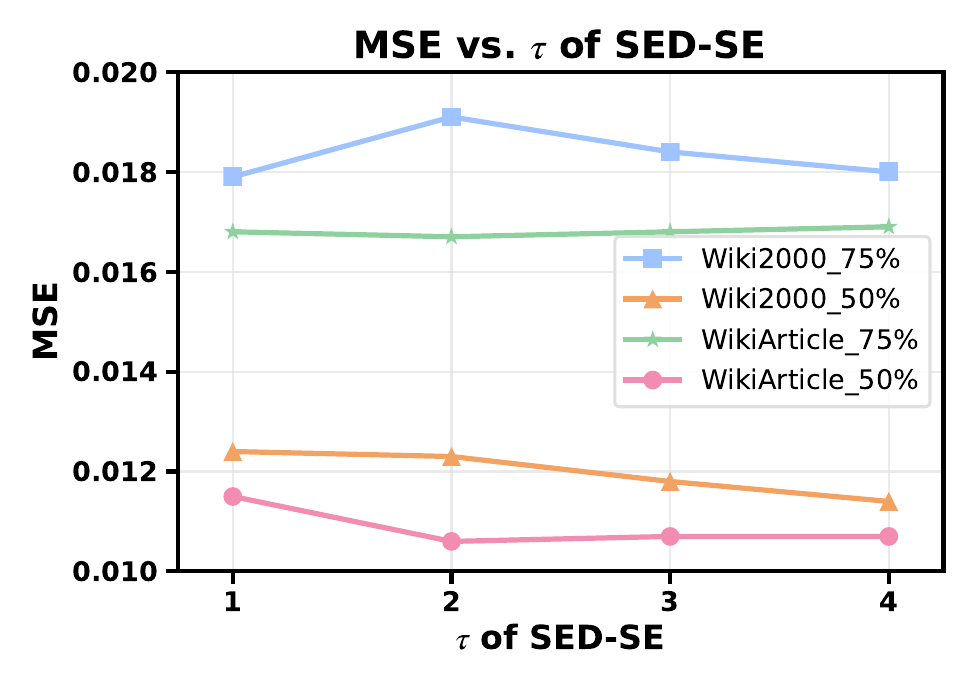}}
  {\includegraphics[width=0.23\linewidth]{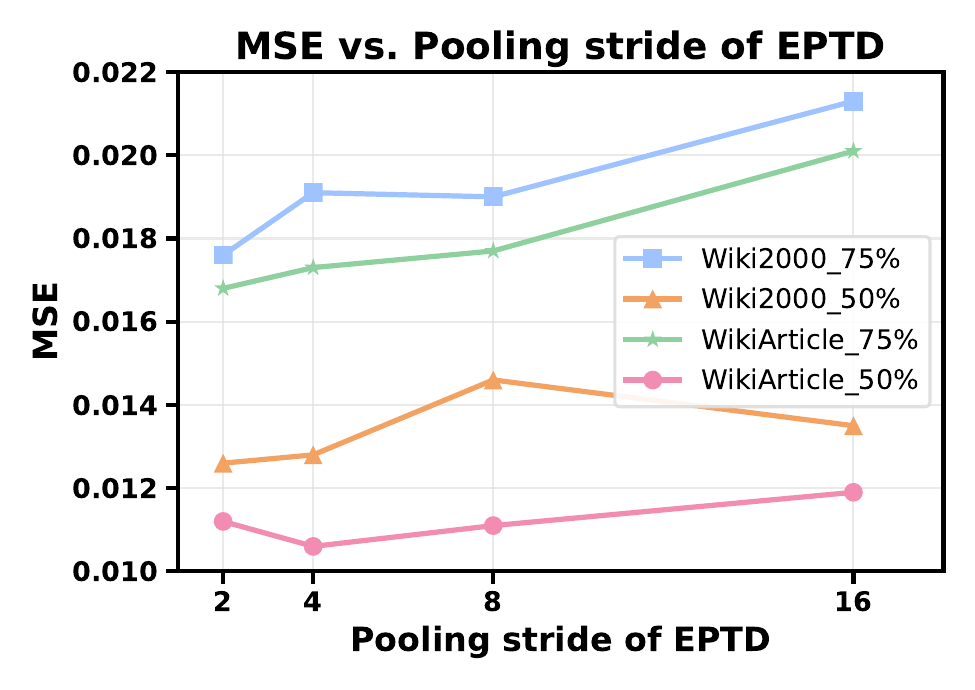}}
  {\includegraphics[width=0.23\linewidth]{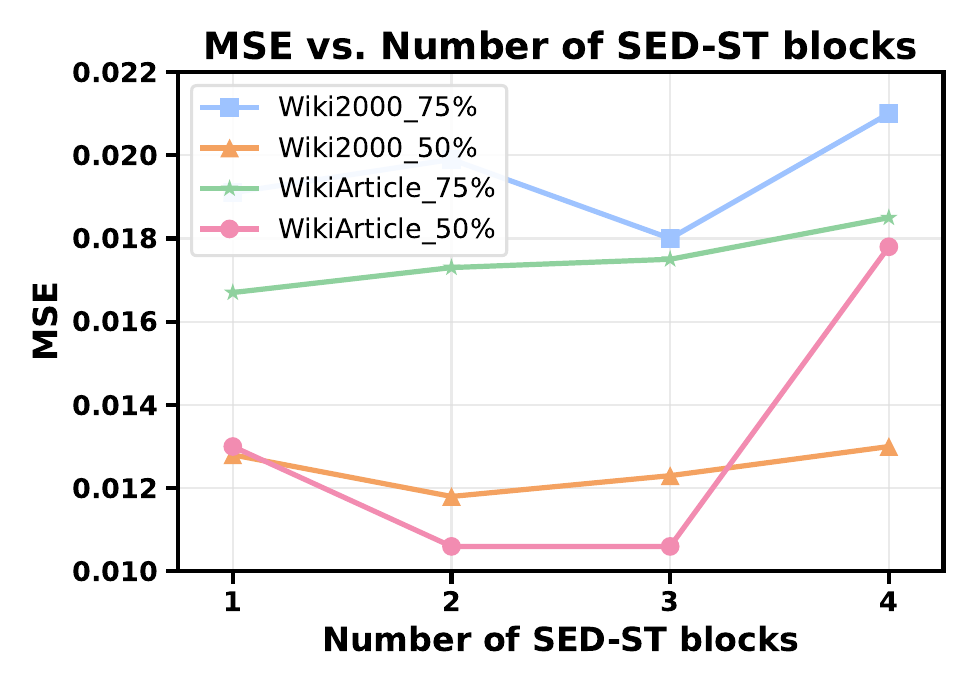}}
  {\includegraphics[width=0.23\linewidth]{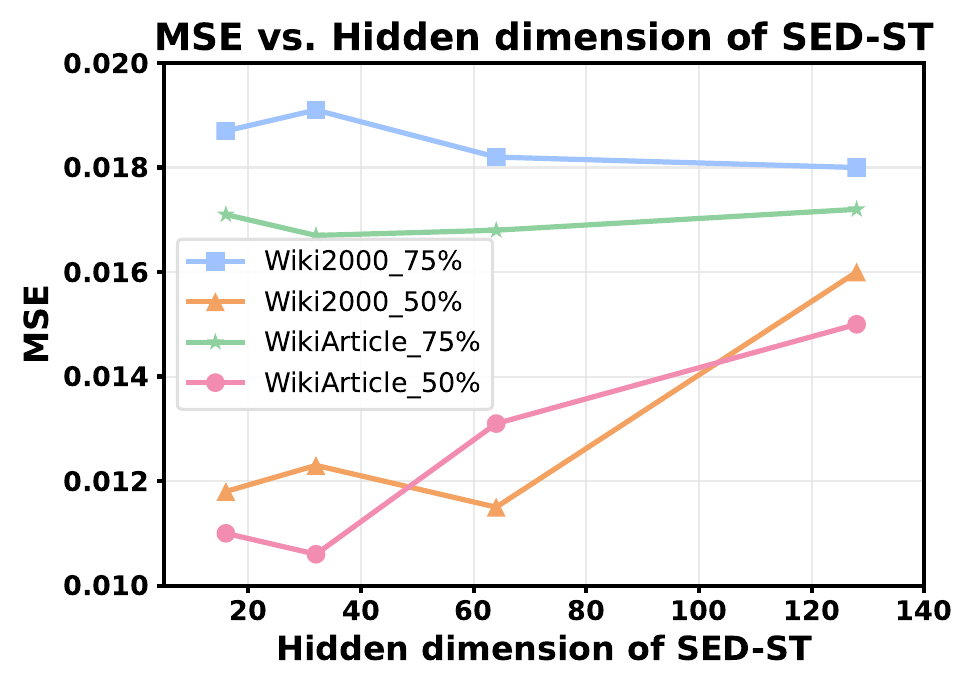}}
  \vspace{-4mm}
  \caption{Hyper–parameter sensitivity analysis of SEDformer evaluated by MSE on time constant $\tau$ of SED-SE, pooling stride of EPTD module, number of SED-ST blocks, hidden dimension of SED-ST.}
  \vspace{-3mm}
  \label{fig:hyper}
\end{figure*}

\subsection{Efficiency Analysis}
\label{sec:efficiency}
As illustrated in Fig.~\ref{fig:efficiency}, we evaluate efficiency on Wiki2000 with a sparsifying rate of \(\mathbf{25\%}\), using the same batch size (16) across all models for fairness. We compare three representative baselines that cover dominant IMTS paradigms: tPatchGNN \cite{tPatchGNN} (patch-level padding), KAFNet \cite{kafnet} (series-level padding), and GraFITi \cite{GraFITi} (bipartite graph without padding). Training time is measured as wall-clock time per epoch on the same GPU stack; our SNN is trained with standard surrogate gradients, enabling straightforward timing. SEDformer uses the fewest parameters and, via event-aligned computation, requires substantially fewer operations than GNN-based methods while remaining comparable to a lightweight ANN baseline. For energy, we follow established digital cost accounting \cite{horowitz20141,lv2024efficient,TS-LIF}: total energy sums memory and arithmetic costs weighted by operation counts, using FLOPs for ANNs and SOPs for SNNs under a 45\,nm technology model. Under this setting, SEDformer reduces theoretical energy by over \(\mathbf{80\times}\) vs.\ tPatchGNN \cite{tPatchGNN}, \(\mathbf{47\times}\) vs.\ GraFITi \cite{GraFITi}, and \(\mathbf{15\times}\) vs.\ KAFNet \cite{kafnet}. These results suggest that event-synchronous spiking can preserve accuracy while markedly improving energy efficiency. Details of the energy accounting are provided in Appendix~\ref{sec:appendixenergy}, and additional results across datasets and sparsifying rates are in Appendix~\ref{app:efficiency}.

\subsection{Spike Trains Visualization}

\label{sec:viz-encoders}
To concretize how spike encoders behave on irregular data, Fig.~\ref{fig:spiketrains} shows spike trains from three representatives on the same synthetic IMTS: a Delta Spike encoder (vanilla LIF)~\cite{lv2024efficient}, a Convolutional Spike encoder (vanilla LIF)~\cite{lv2024efficient}, and our SED-based Spike Encoder (SED-SE with EA-LIF). The two grid-based encoders implicitly assume uniform sampling, so their spikes are quantized to the fixed grid: in the long sparse stretch they miss isolated observations, and around the burst they either emit a few delayed, clustered spikes (Delta) or smear short bursts into wide plateaus with shifted on/off transitions (Convolutional), thus blurring and desynchronizing event timing while offering no gain other than longer, costlier sequences under zero-padding. In contrast, SED-SE updates only on the irregular event grid and modulates leak by the inter-event interval. Long gaps induce strong decay and silence, whereas dense bursts produce concentrated spikes exactly at observation times. This event-synchronous, interval-aware behavior preserves event semantics, exploits sparsity without grid padding, and aligns with the Sparsity–Event Duality that underpins IMTS, explaining the empirical advantages of our model. We describe the procedure for constructing the synthetic IMTS dataset in Appendix \ref{sec:viz-encoders-data}.

\vspace{-2mm}
\subsection{Ablation Study}
\label{sec:ablation}

We conduct ablations across all datasets and Sparsifying Rates to assess the contribution of each component. As summarized in Tab. \ref{tab:ablation_results}, removing any module consistently degrades accuracy, confirming that each stage is functionally necessary. The largest drop arises when the SED-ST blocks are removed, highlighting the importance of modeling temporal irregularity and performing deep representation learning directly on the event-synchronous sequence. Substituting the EA-LIF with the vanilla LIF also hurts performance, indicating that conditioning leakage on inter-event intervals is crucial for irregular data. Likewise, replacing SED-A with standard Self-Attention (SA) leads to consistent declines, showing that our membrane-driven, interval-aware attention better preserves event semantics while exploiting sparsity. These results validate the design choices that align computation with observed events and encode inter-event timing throughout the stack.

\vspace{-2mm}

\subsection{Hyper-parameter Analysis}
\label{sec:hyper}

We study the sensitivity of SEDformer to four key hyper-parameters, results are summarized in Fig.~\ref{fig:hyper}. A moderate leak yields the best accuracy across datasets, whereas too small $\tau$ over-reacts to noise in dense bursts and too large $\tau$ retains stale context across long gaps, confirming the need to balance event persistence and inter-event decay. Smaller strides preserve more event detail and generally improve performance; overly aggressive pooling compresses long gaps but also removes short bursts, degrading accuracy. Increasing the number of backbone blocks helps at first (better temporal representation) and then saturates, with a shallow sweet spot that avoids overfitting and keeps compute modest. Enlarging the width improves capacity up to a point, after which returns diminish and variance grows, suggesting that a mid-range dimension offers a good trade-off between expressiveness and efficiency. Overall, SEDformer is robust in a broad region around these mid-range settings, and the observed trends align with our design intuition: respect events (avoid over-pooling), forget across long gaps (appropriate $\tau$), and favor compact yet sufficiently deep backbones.

\vspace{-1mm}

\section{Conclusion and Future Work}
We present SEDformer, a Spiking Transformer that models web telemetry irregular multivariate time series (IMTS) by aligning computation to observed events and conditioning dynamics on inter–event intervals (which is identified and summarized as the Sparsity–Event Duality, SED). The proposed SEDformer integrates an SED-based Spike Encoder, Event–Preserving Temporal Downsampling module, and a stack of SED-based Spike Transformer blocks, which together preserve event semantics while exploiting sparsity for efficiency. Experiments on public web telemetry datasets demonstrate that this event-synchronous, interval-conditioned approach delivers state-of-the-art accuracy with reduced compute and energy, while our analyses show that grid-padding pipelines and relational recastings tend to dilute short bursts by inflating sequences and disrupting local temporal continuity. 

Looking ahead, we plan to extend the model with probabilistic forecasting and calibration for risk-aware operations, study online learning for streaming settings with distribution shift, and explore deployment on low-power hardware and neuromorphic backends to further translate spike sparsity into practical efficiency gains.

\clearpage
\bibliographystyle{ACM-Reference-Format}
\bibliography{main}

\clearpage
\appendix 
\begingroup
\centering
{\LARGE\bfseries Appendix\par}
\endgroup

\section{Additional Details on Dataset Construction}
\label{app:datasets}

\noindent\textbf{Corpora.}
We use two public telemetry IMTS benchmarks that originate from web-traffic telemetry: Wiki2000 \cite{Wike2000} and WikiArticle \cite{WikeArticle}.
For Wiki2000 we follow TFB \cite{qiu2024tfb} and keep the first 200 variates (daily, 2015-07-01--2017-09-10).
For WikiArticle we load \texttt{train\_1.csv} and keep the first 300 variates (daily, 2015-07-01--2016-12-31).
Choosing the provider's first $200/300$ series ensures parity of training cost and avoids cherry-picking bias.

\noindent\textbf{Protocol.}
Raw series contain long gaps and sporadic outliers. We first create a clean reference trajectory per variate on the daily grid, then induce irregular sampling via independent masking (MCAR) \cite{jager2021benchmark}, following common IMTS benchmarking practice \cite{SUSTeR,du2024tsibench,Latent-ODE,gpvae2020}.
We evaluate in a rolling multi-step setting with a 90-day history and 30-day horizon; sparsifying rates $r\!\in\!\{0.25,0.50,0.75\}$ use a fixed seed across methods.
The complete cleaning--sparsifying pipeline is summarized in Alg.~\ref{alg:wikiclean}.

\begingroup\small
\begin{algorithm}[!ht]
\caption{Telemetry benchmarks: Clean $\rightarrow$ MCAR sparsify (per variate)}
\label{alg:wikiclean}
\begin{algorithmic}[1]
\Require $x\!\in\!\mathbb{R}^{T}$ (daily); window $w$; gap cap $L_{\max}$; outlier $\tau$; rate $r$; seed $s$
\Ensure $\bar{x}$ on irregular times \textbf{or} $(\tilde{x},m)$ with $m\!\in\!\{0,1\}^{T}$
\State $\tilde{x}\!\gets x$
\State \emph{Short gaps} ($\le L_{\max}$): quadratic Lagrange; else boundary bridge
\State \emph{Outliers}: with median $\mu$, MAD $\sigma$, if $|\tilde{x}_t-\mu|/\sigma>\tau$ then replace by local median (size $w$)
\State Spline-impute any remaining long gaps
\State RNG $\gets s$; sample $m_t\sim{\rm Bernoulli}(1-r)$ i.i.d.
\State \textbf{return} $\bar{x}=\{(t,\tilde{x}_t):m_t=1\}$ \textbf{or} $(\tilde{x},m)$
\end{algorithmic}
\end{algorithm}
\endgroup

\section{Baselines}
\label{app:baselines}
\begin{itemize}[leftmargin=1.1em]
  \item \textbf{Neural ODE}~\citep{Neural-ODE}: Views hidden dynamics as a continuous-time ODE and integrates forward with an ODE solver; forecasts are obtained by decoding states at queried times.
  \item \textbf{Latent ODE}~\citep{Latent-ODE}: Encodes an irregular history into an initial latent state, evolves it with a neural ODE, and decodes values at arbitrary future timestamps, enabling fully continuous-time forecasting.
  \item \textbf{GRU-ODE}~\citep{GRU-ODE}: Maintains a GRU state that decays according to elapsed time between events and is updated only at observation times, combining event-driven updates with continuous-time evolution.
  \item \textbf{GRU-D}~\citep{Che2018}: A GRU with learnable time decay and feature-wise imputation; uses masks and time gaps to modulate hidden updates under missingness and irregular intervals.
  \item \textbf{SeFT}~\citep{seft}: Treats samples as an unordered set of $(\text{time},\text{value},\text{variable})$ tuples and aggregates with permutation-invariant set functions (Deep Sets) for irregular inputs.
  \item \textbf{RainDrop}~\citep{RainDrop}: Builds a time-aware graph over observations and applies message passing with temporal attention to fuse information across irregular stamps and variables.
  \item \textbf{Warpformer}~\citep{Warpformer}: Learns a timestamp/value warping to align multi-scale patterns before Transformer encoding, improving robustness to nonuniform sampling.
  \item \textbf{mTAND}~\citep{mTAND}: Learns continuous-time embeddings via attention over reference points; for forecasting, replaces interpolation queries by future queries to obtain predictions at desired times.
  \item \textbf{CRU}~\citep{CRU}: Uses continuous recurrent units derived from linear SDE dynamics with continuous--discrete filtering, providing principled uncertainty handling for irregular streams.
  \item \textbf{Neural Flows}~\citep{Neural-Flows}: Reparameterizes neural-ODE trajectories as continuous normalizing flows to improve stability/efficiency while retaining continuous-time flexibility.
  \item \textbf{tPatchGNN}~\citep{tPatchGNN}: Forms transformable temporal patches per series, models intra-patch patterns with a Transformer, and uses time-adaptive GNNs to capture inter-variable relations.
  \item \textbf{GraFITi}~\citep{GraFITi}: Constructs a time- and variable-aware graph directly on unaligned observations; stacked diffusion/attention operators propagate signals to infer future values without pre-alignment.
  \item \textbf{TimeCHEAT}~\citep{TimeCHEAT}: Enforces channel harmony by combining channel-dependent graph convolutions within patches with channel-independent Transformers across patches, balancing specialization and sharing.
  \item \textbf{KAFNet}~\citep{kafnet}: Revitalizes canonical pre-alignment via a lightweight kernel-alignment/feature pre-alignment module that maps irregular series to a unified temporal basis, enabling simple yet competitive forecasting on unaligned data.
  \item \textbf{HyperIMTS}~\citep{HyperIMTS}: Models IMTS as a hypergraph where observations are nodes and temporal/variable hyperedges capture higher-order dependencies; message passing occurs directly on irregular event sets.
\end{itemize}

\section{Hyper-parameters Search Space}
\label{app:hyperparameter}

We sweep the following hyperparameters for our model:
\begin{itemize}
    \item $\tau$ in SED-SE: $\tau \in \{1,2,3,4\}$.
    \item Pooling stride of EPTD: $s \in \{2,4,8,16\}$.
    \item Number of SED-ST blocks: $L \in \{1,2,3,4\}$.
    \item Hidden dimension in SED-ST: $d \in \{16, 32,64,128\}$.
\end{itemize}

\section{Theoretical Energy Consumption Calculation}
\label{sec:appendixenergy}
We follow standard digital cost accounting \cite{lv2024efficient,TS-LIF} with Horowitz's 45\,nm energy model \cite{horowitz20141}. 
For a model with layers $\mathcal{L}$, the theoretical energy is
\[
E_{\text{tot}}
=\sum_{\ell\in\mathcal{L}}\!\Big(
N^{(\ell)}_{\rm MAC}\,e_{\rm MAC}
+N^{(\ell)}_{\rm add}\,e_{\rm add}
+N^{(\ell)}_{\rm rd}\,e_{\rm rd}
+N^{(\ell)}_{\rm wr}\,e_{\rm wr}
\Big),
\]
where $e_{\rm MAC}, e_{\rm add}$ are operation energies and $e_{\rm rd}, e_{\rm wr}$ are SRAM read/write energies under 45\,nm where $e_{\rm MAC} = 4.6 \, \text{pJ}$ and $e_{\rm add} = 0.9 \, \text{pJ}$.

\noindent\textbf{ANN layers.}
For a dense linear (or $1{\times}1$ conv) with $d_{\rm in}\!\times\! d_{\rm out}$ and effective sequence length $T_{\rm eff}$, 
$N_{\rm MAC}=d_{\rm in}d_{\rm out}T_{\rm eff}$ and $N_{\rm add}\!\approx\! d_{\rm out}(d_{\rm in}{-}1)T_{\rm eff}$; memory terms account for parameter fetches and activations. 
With MCAR masking, grid-based models still process the full grid so $T_{\rm eff}=T$, whereas true event-driven models use $T_{\rm eff}=|\{t_k:m_k{=}1\}|$.

\noindent\textbf{SNN (event-driven).}
For a spiking layer with fan-in $d_{\rm in}$ and spike count $S_\ell$, a spike-operation (SOP) bundles an accumulate, a compare/reset, and a few SRAM accesses:
\[
E^{(\ell)}_{\rm SNN}
= S_\ell\,(e_{\rm acc}{+}e_{\rm cmp}{+}e_{\rm rd}{+}e_{\rm wr}) \;+\; E_{\rm params}^{(\ell)},
\]
with $S_\ell \approx \rho_\ell\,|\{t_k:m_k{=}1\}|\,d_{\rm out}$ given firing rate $\rho_\ell$. 
SEDformer is event-aligned (counts scale with observed events), so its OPs/SOPs and energy shrink with sparsity. 
Under the same 45\,nm assumptions, this yields the large theoretical savings reported in Fig.~\ref{fig:efficiency} (e.g., $>80\times$ vs.\ tPatchGNN \citep{tPatchGNN}, $47\times$ vs.\ GraFITi \citep{GraFITi}, and $\mathbf{15\times}$ vs.\ KAFNet \citep{kafnet}).

\section{Additional Efficiency Analysis}
\label{app:efficiency}

\begin{figure}[!ht]
  \captionsetup[sub]{skip=0.1pt}
  \centering
  \subfloat[Wiki2000 50\%]{
    \includegraphics[width=0.49\linewidth]{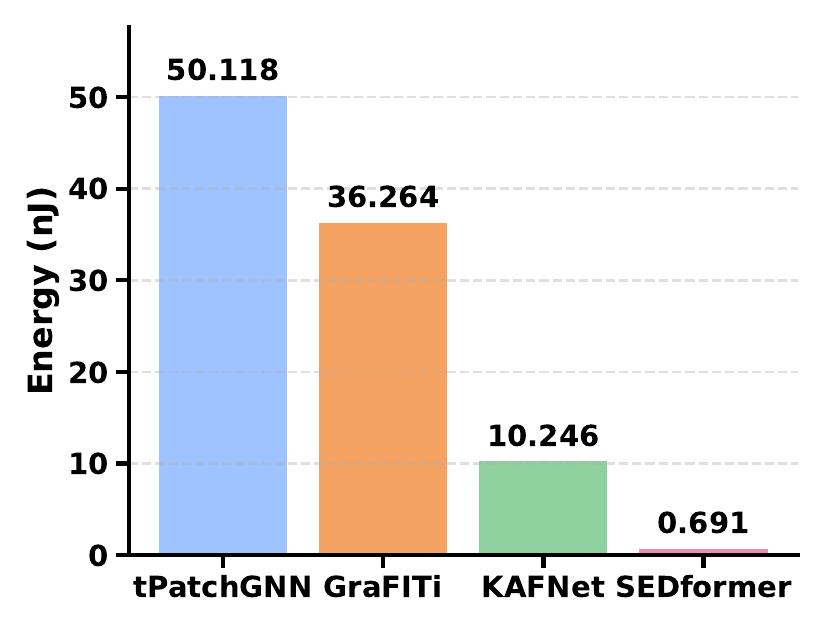}}
  \hfill
  \subfloat[Wiki2000 75\%]{
    \includegraphics[width=0.49\linewidth]{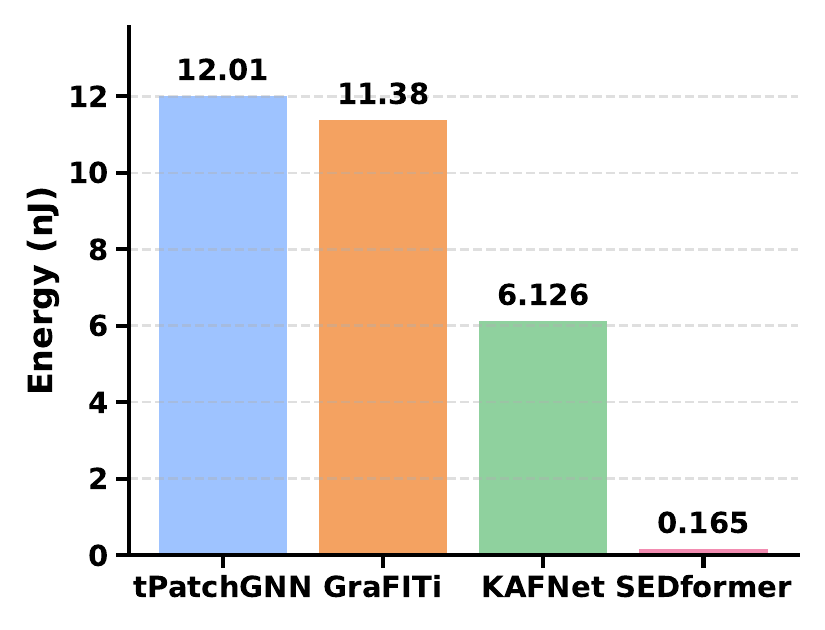}}
  \par\smallskip
  \subfloat[WikiArticle 50\%]{
    \includegraphics[width=0.49\linewidth]{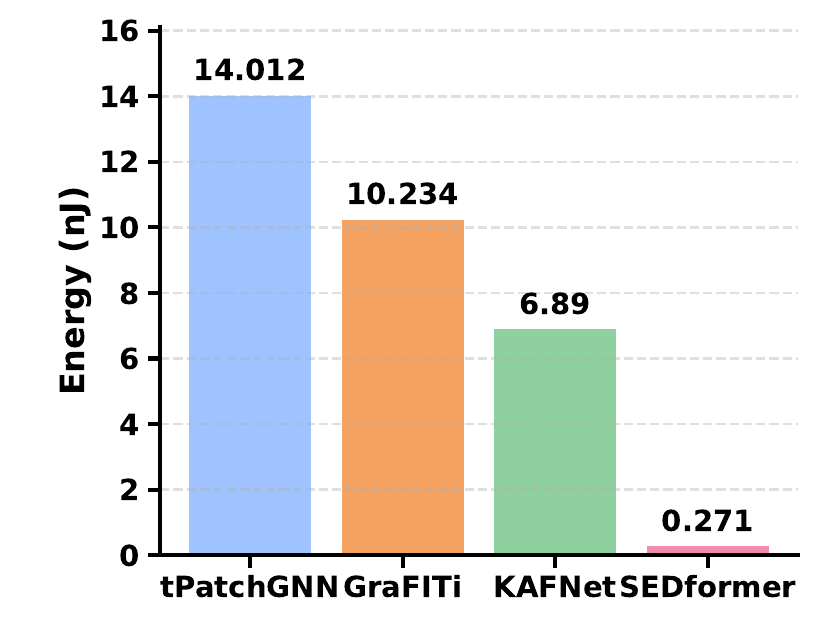}}
  \hfill
  \subfloat[WikiArticle 75\%]{
    \includegraphics[width=0.49\linewidth]{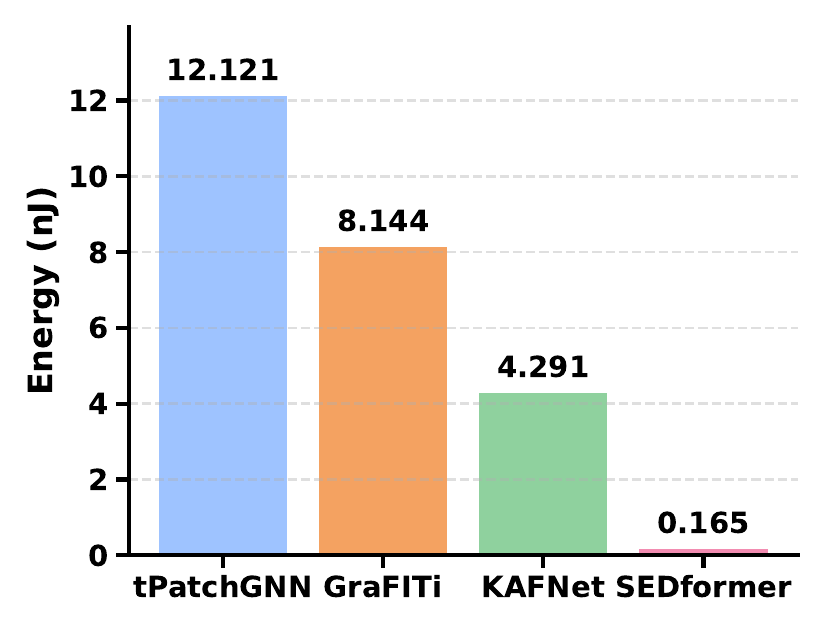}}
  \caption{\textbf{Extended energy comparison under higher sparsity.} Theoretical energy is computed using the same digital cost model as Sec.~\ref{sec:efficiency} (Horowitz 45\,nm), with SOPs for SEDformer and FLOPs for ANN baselines. SEDformer retains a consistent advantage across datasets and sparsifying rates, and the margin tends to increase as sparsity grows, highlighting the benefit of event-synchronous computation.}
  \label{fig:appendix-energy}
\end{figure}

To substantiate the efficiency claims in Sec.~\ref{sec:efficiency}, we report additional results on higher sparsifying rates in Fig.~\ref{fig:appendix-energy}. We follow the same accounting protocol as the main text (45\,nm digital energy model with Horowitz costs; SOPs for SNNs and FLOPs for ANNs; identical GPU, batch size \(=16\), and software stack). Across both datasets and masking levels, SEDformer consistently achieves markedly lower theoretical energy than representative ANN baselines. The gap generally widens as sparsity increases, reflecting that event-synchronous computation scales with the number of pooled events rather than the length of a padded grid, thereby avoiding work on non-informative steps.

\section{Spike-Encoder Visualization Dataset}
\label{sec:viz-encoders-data}

To visualize spike encoders (Sec.~\ref{sec:viz-encoders}), we synthesize an irregular multivariate time series on a single channel. 
Let $f(t)$ be a smooth baseline with two bursts; we sample (i) a \emph{sparse} phase on $[0,60]$ with $n_s$ points and (ii) a \emph{dense} phase on $(60,T]$ with $n_d$ points ($T{=}100$, seed fixed). 
Observed values are $x_k=f(t_k)+\epsilon_k$ with small Gaussian noise. 
We also build a regular grid $\{t_k^{\rm reg}\}_{k=1}^{N}$ and its values $x_k^{\rm reg}=f(t_k^{\rm reg})+\epsilon_k$ to run grid-based encoders.

\noindent\textbf{Encoders.}
(i) \emph{Delta (regular)}: $s_k^{\Delta}=\mathbf{1}\big(|x_k^{\rm reg}-x_{k-1}^{\rm reg}|\ge \theta\big)$. 
(ii) \emph{Convolutional (regular)}: $y=\kappa * x^{\rm reg}$ with a short kernel $\kappa$; $s_k^{\rm conv}=\mathbf{1}\big( (y_k-\bar y)/{\rm std}(y) \ge \tau_c \big)$. 
(iii) \emph{SED-SE (irregular, EA-LIF)}: for irregular stamps $\{t_k\}$,
\(
\beta_k=\exp\!\big(-\tfrac{t_k-t_{k-1}}{\tau}\big),\; 
m_k=\beta_k v_{k-1}+(1-\beta_k)\,\gamma(x_k-\theta),\;
s_k=\mathbf{1}(m_k\ge v_{\rm th}),\;
v_k=m_k-v_{\rm th}\,s_k .
\)
We re-implement the Delta and the Convolutional spike encoders from \citep{lv2024efficient}. Grid encoders quantize to the fixed step, which blurs timing under sparsity; SED-SE fires exactly at observation times and modulates leak by inter-event intervals, yielding event-synchronous behavior.

\section{Discussions and Limitations}
\label{appendix:limitations}
Our proposed SEDformer broadens the IMTS forecasting toolkit by demonstrating that spiking neural networks can deliver energy-efficient, robust, and accurate predictions. We hope it offers a fresh perspective to the data mining and time-series community and stimulates further research on event-native modeling of irregular telemetry streams. Moreover, SEDformer's event-synchronous computation aligns naturally with bursty, asynchronous telemetry (e.g., page views, service metrics, sensors, and device logs), preserving the information of real observation events rather than padded grid steps. Our event-synchronous spiking design with interval-aware leakage also has limitations. First, its practical efficiency hinges on kernels and runtimes that exploit irregular timestamps; on dense tensor backends the sparsity benefits may be underutilized. Second, our training and evaluation primarily assume MCAR-style sparsity; under MNAR or structured censoring the model can behave suboptimally without retraining or auxiliary propensity/imputation mechanisms. Third, spike thresholds, leak constants and interval scalings couple to dataset-specific sampling rates, so cross-dataset transfer may require careful retuning to avoid under-/over-firing. Finally, our energy accounting follows standard 45\,nm digital cost models and reports theoretical, not measured, consumption; absolute numbers will vary with hardware and low-level implementations even if the relative trends remain.

\end{document}